\documentclass{article}

     \PassOptionsToPackage{numbers, compress}{natbib}
\usepackage[preprint]{neurips_2026}

\usepackage[utf8]{inputenc} % allow utf-8 input
\usepackage[T1]{fontenc}    % use 8-bit T1 fonts
\usepackage{hyperref}       % hyperlinks
\usepackage{url}            % simple URL typesetting
\usepackage{booktabs}       % professional-quality tables
\usepackage{amsfonts}       % blackboard math symbols
\usepackage{nicefrac}       % compact symbols for 1/2, etc.
\usepackage{microtype}      % microtypography
\usepackage{xcolor}         % colors
\usepackage{enumitem}       % enumerate with letters

\DeclareMathAlphabet{\mathcal}{OMS}{cmsy}{m}{n} %fix what mathptmx does to \mathcal
\usepackage{bbold}
\usepackage{amsmath}
\usepackage{amssymb}
\usepackage{amsthm}
\newtheorem{theorem}{Theorem}
\newtheorem{lemma}{Lemma}
\newtheorem{corollary}{Corollary}
\newtheorem{proposition}{Proposition}

\usepackage{algorithm}
\usepackage[noend]{algpseudocode} 
\usepackage[pdftex]{graphicx}
\usepackage{hyperref}
\usepackage[capitalise, nameinlink]{cleveref} % must load AFTER hyperref
\usepackage[title]{appendix}

\usepackage{silence}
\usepackage{acronym}
\newacro{KAN}[KAN]{Kolmogorov-Arnold Network}
\newacro{KAPLAN}[KAPLAN]{KAPLAN}
\newacro{PH}[PH]{proportional hazards}
\newacro{NPH}[NPH]{non-proportional hazards}
\newacro{HR}[HR]{hazard regression}
\newacro{PMF}[PMF]{probability mass function}
\newacro{AFT}[AFT]{accelerated failure time}
\newacro{FPM}[FPM]{flexible parametric model}
\newacro{GAM}[GAM]{generalised additive model}
\newacro{DGP}[DGP]{data-generating process}
\newacro{KAPLAN-HR}[KAPLAN-HR]{KAPLAN hazard regression}
\newacro{MLP}[MLP]{multi-layer perceptron}
\newacro{NLL}[NLL]{negative log-likelihood}
\newacro{DNN}[DNN]{deep neural network}
\newacro{METABRIC}[METABRIC]{Molecular Taxonomy of Breast Cancer International Consortium}
\newacro{RotGBSG}[RotGBSG]{Rotterdam tumor bank and German Breast Cancer Study Group}
\newacro{NWTCO}[NWTCO]{National Wilms' Tumor Study}
\newacro{FLCHAIN}[FLCHAIN]{Assay Of Serum Free Light Chain}
\newacro{SUPPORT}[SUPPORT]{Study to Understand Prognoses Preferences Outcomes and Risks of Treatment}
\newacro{MIMIC-III}[MIMIC-III]{Medical Information Mart for Intensive Care III}
\newacro{C-TD}[C-TD]{time-dependent C-index}
\newacro{IBS}[IBS]{integrated Brier score}
\newacro{ICI}[ICI]{integrated calibration index}
\newacro{D-CAL}[D-CAL]{D-calibration}
\newacro{MC}[MC]{Monte Carlo}
\newacro{MLE}[MLE]{maximum-likelihood estimator}
\newacro{HARE}[HARE]{HARE}
\newacro{ISE}[ISE]{integrated squared error}
\newacro{IQR}[IQR]{inter-quartile range}
\newacro{DSM}[DSM]{deep survival machines}

\title{KAPLAN: Kolmogorov-Arnold Prognostic Learnable Activation Networks for Survival Analysis}

\author{%
  Stelios Boulitsakis Logothetis \\
  University of Cambridge \\
  \texttt{sb2690@cam.ac.uk} \\
  \And
  Angela Wood \\
  University of Cambridge \\
  \texttt{amw76@cam.ac.uk} \\
  \And
  Pietro Li\`{o}\\
  University of Cambridge \\
  \texttt{pl219@cam.ac.uk} 
}

\begin{document}

\maketitle

\begin{abstract}
    Survival analysis aims to model how covariates and time jointly shape the time-to-event distribution under right censoring. Classical methods such as the Cox model and generalised additive models (GAMs) require interactions and time-varying effects to be manually specified, which is increasingly impractical on rich clinical datasets. We introduce KAPLAN-HR, a B-spline Kolmogorov-Arnold Network (KAN) for nonparametric estimation of the conditional hazard as a joint function of covariates and time. A single-layer KAPLAN-HR model recovers a GAM, while deeper architectures capture interactions and time-varying effects through composition. We establish a convergence rate for the nonparametric KAN hazard estimator that depends only on the smoothness of the underlying KAN representation and not on the covariate dimension, thereby mitigating the curse of dimensionality for KAN-representable targets. In evaluations over six clinical benchmark datasets, KAPLAN-HR matches or exceeds the predictive performance of established statistical and deep learning survival methods.
\end{abstract}

\section{Introduction}
\acresetall
\acused{HARE}
\acused{KAPLAN-HR}

Time-to-event data is ubiquitous in biostatistics, engineering, and business, but analysing it is complicated because the events of interest are often only partially observed. Patients may leave a study before their disease progresses, or a machine may be replaced before it fails, so only a lower bound on the true event time is ever observed for these subjects. Survival analysis addresses this by modelling the time-to-event distribution under this incomplete observation, called right censoring.

The Cox \ac{PH} model is the de facto standard for survival regression \cite{coxRegressionModelsLifeTables1972}, but it requires careful manual work to accommodate interactions between features (covariates) and restricts their relationship with time. Among the more flexible alternatives, spline-based approaches for modelling the hazard have been studied for over three decades. For example, \acp{GAM} with a survival likelihood fit a spline for each covariate's effect on the hazard \cite{hastieGeneralizedAdditiveModels1995, woodGeneralizedAdditiveModels2025}. \ac{HARE} models the log-hazard as linear splines and selects main effects and pairwise interaction terms by stepwise search \cite{kooperbergHazardRegression1995}. However, these models are structurally limited in the way they represent interactions, as \acp{GAM} only include them when interaction terms are specified manually, and \ac{HARE} cannot represent three-way or higher interactions. 

This limitation is increasingly consequential, particularly in medicine and biostatistics, where modern datasets such as UK Biobank \cite{sudlowUKBiobankOpen2015} and MIMIC \cite{johnsonMIMICIIIClinicalDatabase2016} are far richer than those available before, motivating methods that capture interactions and time-varying effects without manual specification. Deep learning approaches \cite{wiegrebeDeepLearningSurvival2024, chenIntroductionDeepSurvival2024} relax classical models' structural restrictions and perform particularly well on high-dimensional data such as imaging, but depart from the additive structure that makes classical models simple to analyse and interpret.

\acp{KAN} \cite{liuKANKolmogorovarnoldNetworks2025} offer a middle ground. Each edge of a \ac{KAN} applies a learnable univariate function and each node sums its inputs. When the edge functions are splines, a one-layer \ac{KAN} reduces to a \ac{GAM}-like additive model, while deeper architectures introduce interactions through composition. Recent work has applied \acp{KAN} to survival analysis via different parameterisations, including proportional hazards \cite{knottenbeltCoxKANKolmogorovArnoldNetworks2025, cheng2025extendingcoxproportionalhazards}, accelerated failure time \cite{jose2025kanaftinterpretablenonlinearsurvival}, and nonparametric hazard regression \cite{mastroleo2026survkanfullyparametricsurvival}. These studies report encouraging results, but their focus is primarily on empirical performance and on interpretability via edge visualisation and symbolic regression, and none have yet provided theoretical analysis. Analysing the statistical properties of \ac{KAN}-based survival estimators is therefore of substantive interest.

We introduce \ac{KAPLAN-HR}, a B-spline \ac{KAN} model that estimates the conditional hazard as a joint function of covariates and time. A single-layer \ac{KAPLAN-HR} recovers a B-spline \ac{GAM} under the \ac{PH} restriction, and deeper architectures capture covariate interactions and time-varying effects through composition. We make the following contributions: (i) We propose \ac{KAPLAN-HR} as a bridge between additive spline-based hazard models and deep survival models, where depth controls the level of structural relaxation. (ii) We establish a convergence rate of the proposed nonparametric estimator to the true \ac{KAN}-representable log-hazard that is independent of the input dimension. (iii) We empirically evaluate \ac{KAPLAN-HR} on six clinical benchmark datasets, where it matches or exceeds established statistical and deep learning survival models.

\begin{figure}[t]
    \centering
    \includegraphics[width=\linewidth]{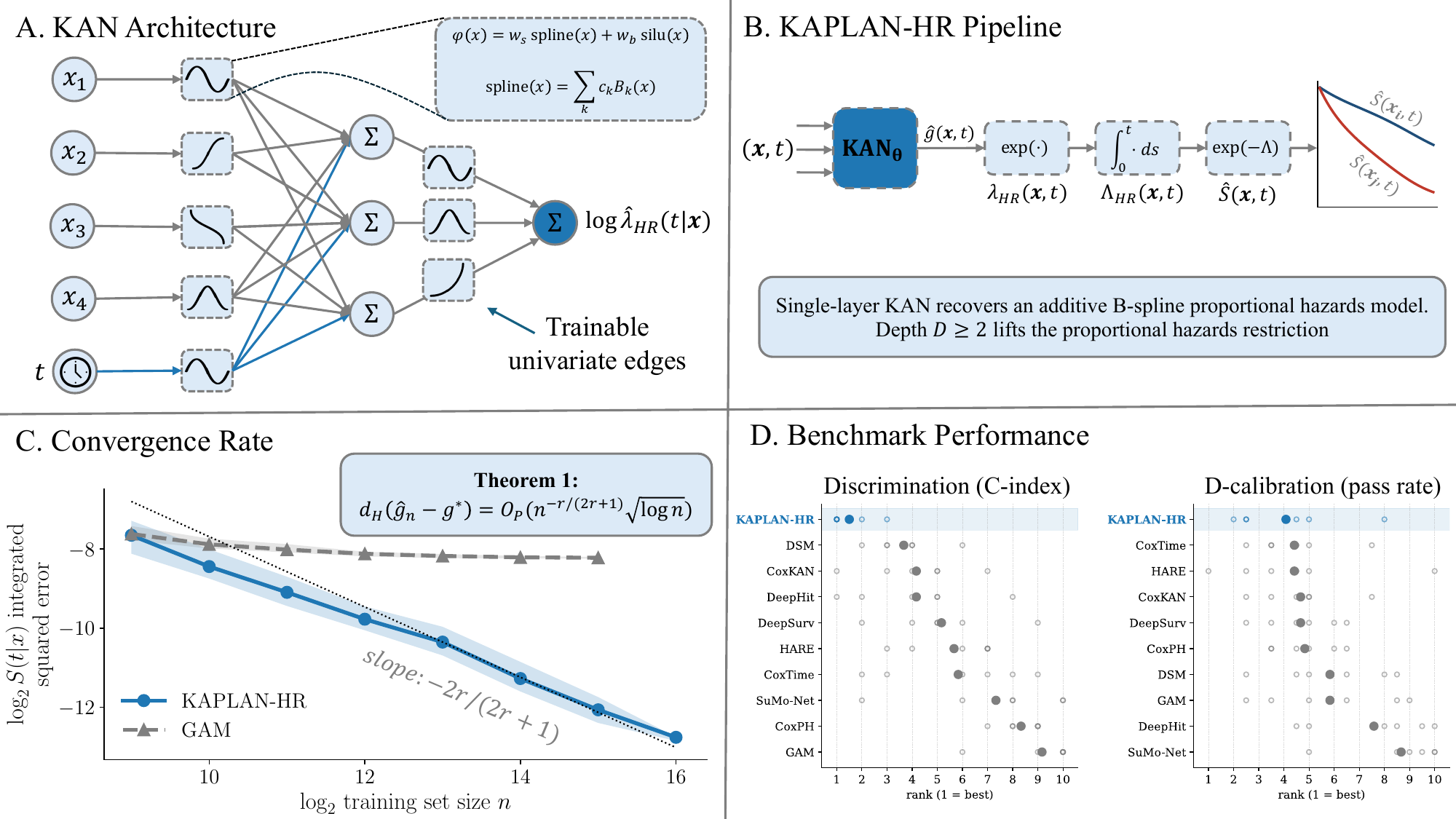}
    \caption{
        \textbf{Overview of \ac{KAPLAN-HR}.}
        \textbf{(A)} \acp{KAN} compose layers of learnable univariate functions. We propose using a \ac{KAN} to estimate the log-hazard as a joint function of covariates and time.
        \textbf{(B)} Numerical integration transforms the predicted log-hazard into predicted survival curves. A single-layer \ac{KAPLAN-HR} recovers a \ac{GAM}-like proportional hazards additive B-spline model, while deeper models capture covariate interactions and time-varying effects through composition.
        \textbf{(C)} We establish a nonparametric convergence rate of $O_P(n^{-r/(2r+1)}\sqrt{\log n})$ for the \ac{KAPLAN-HR} estimator. Simulation studies corroborate this rate and demonstrate that \ac{KAPLAN-HR} captures interaction relationships that additive models cannot.
        \textbf{(D)} On six clinical benchmark datasets, \ac{KAPLAN-HR} matches or exceeds established statistical and deep learning survival models in terms of discrimination, survival accuracy, and calibration.
    }
    \label{fig:abstract}
\end{figure}

\section{Preliminaries}
\label{sec:preliminaries}
Denote by $\mathcal{D}$ a time-to-event dataset $\{(\mathbf{x}_i, Y_i, \Delta_i)\}_{i=1}^n$, comprising $n$ i.i.d. observations of continuous covariates $\mathbf{x}_i \in [0, 1]^d$, observed times $Y_i \in [0, 1]$, and event indicators $\Delta_i \in \{0, 1\}$. We assume the covariates are scaled to the unit cube, and the times are scaled by a finite study horizon $\tau_{\max} \in (0, \infty)$, so $Y_i = 1$ corresponds to subject $i$ surviving to or beyond $\tau_{\max}$. Let $T_i$ denote the (possibly unobserved) time the event of interest occurred, and $C_i$ the censoring time. Under right censoring, we only observe the earlier of the two: $Y_i := \min\{T_i, C_i\}$ and $\Delta_i := \mathbb{1}\{T_i \leq C_i\}$. 

We adopt the common assumption that $T$ and $C$ are independent conditional on $X$. We consider the censoring distribution $\mathbb{P}_{C \mid \mathbf{x}}$ to combine a continuous density $f_C(\cdot \mid \mathbf{x})$ on $[0, 1)$ with a point mass $\alpha(\mathbf{x}) \in [0, 1]$ at $t = 1$ (with $f_C \geq 0$ and $\int_0^1 f_C(s \mid \mathbf{x})\,ds + \alpha(\mathbf{x}) = 1$). This accommodates administrative censoring, where every subject still at risk at $\tau_{\max}$ is censored simultaneously, which is a standard feature of clinical survival datasets.

\paragraph{Hazard model} The ubiquitous Cox \ac{PH} \cite{coxRegressionModelsLifeTables1972} model is defined in terms of the hazard function $\lambda (t \mid \mathbf{x}) = -\frac{d}{dt} \ln S(t \mid \mathbf{x})$, which it decomposes into a feature-invariant baseline hazard function $\lambda_0(t)$ and a time-invariant log-partial hazard function $g(\mathbf{x}; \boldsymbol{\theta}) : \lambda_\text{PH}(t \mid \mathbf{x}) = \lambda_0(t) \exp\!\left(g(\mathbf{x}; \boldsymbol{\theta})\right)$. This constrains the predicted hazard $\lambda_\text{PH}(t \mid \mathbf{x})$ to the same shape as the baseline $\lambda_0(t)$ and the covariate effects to be constant across time. We opt for a more flexible \acf{HR} model where $g(\cdot)$ expresses both the overall relationship with time and the covariate effects, similar to \cite{kooperbergHazardRegression1995}:
\begin{equation}
    \lambda_\text{HR}(t \mid \mathbf{x}) = \exp\!\left(g(\mathbf{x}, t; \boldsymbol{\theta})\right)
    \label{eq:hr_parameterisation}
\end{equation}
We assume the true log-hazard $g^{*}$ is bounded, $\|g^{*}\|_\infty \leq M_0 < \infty$, $M_0 \geq 1$.

This model is fitted by minimising the \ac{NLL} of the observed data, which we express in terms of hazards as:
\begin{equation}
    \mathcal{L}_\text{HR}(\boldsymbol{\theta}) = -\frac{1}{n}\sum_{i=1}^{n} \left[\Delta_i \log \lambda_\text{HR}(Y_i \mid \mathbf{x}_i) - \Lambda_\text{HR}(Y_i \mid \mathbf{x}_i)\right]
    \label{eq:loss_hr}
\end{equation}
The cumulative hazard then defined $\Lambda_{\text{HR}}(t \mid \mathbf{x}) = \int_0^t \lambda_{\text{HR}}(s \mid \mathbf{x})\,ds$ and the survival $S(t \mid \mathbf{x}) = \exp(-\Lambda_{\text{HR}}(t \mid \mathbf{x}))$. 

\section{Methodology}
\label{sec:methodology}
\ac{KAPLAN-HR} represents the learnable function $g(\cdot)$ in \cref{eq:hr_parameterisation} as a \ac{KAN}. \acp{KAN} are neural networks in which the edges carry trainable univariate functions and the nodes sum their inputs \cite{liuKANKolmogorovarnoldNetworks2025}. Each edge is parameterised as the weighted sum of a B-spline and a residual SiLU term (\cref{fig:abstract} (A)). 

Consider a \ac{KAN} consisting of $D$ layers $\Phi_\ell$ having $d_\ell$ nodes, and trainable edges $\varphi_{\ell, i, j}$ that are aggregated by the nodes via summation:
\begin{equation}
    z_j^{(\ell)} = \left[\Phi_\ell\left(\mathbf{z}^{(\ell-1)}\right)\right]_j = \sum_{i=1}^{d_{\ell-1}} \varphi_{\ell, i, j}\!\left(z_i^{(\ell-1)}\right), \qquad \ell = 1, \ldots, D, \quad j \in \{1, \ldots, d_\ell\}
    \label{eq:kan_layer}
\end{equation}
The full network is the composition $\mathrm{KAN}_{\boldsymbol{\theta}}(\mathbf{z}) = (\Phi_D \circ \cdots \circ \Phi_1)(\mathbf{z}) = z_1^{(D)}$.  We refer readers to \citet{noorizadegan2026practitionersguidekolmogorovarnoldnetworks} for further details about the architecture, training, and alternative choices of basis functions for \acp{KAN}. 

Let $\mathrm{KAN}_{\boldsymbol{\theta}}$ be defined as above, with input $\mathbf{z}^{(0)} = (\mathbf{x}, t) \in \mathcal{Z} := [0, 1]^{d_0}$ where $d_0 = d + 1$ (covariate dimension $d$ plus time). We propose the \ac{KAPLAN-HR} survival model:
\begin{equation}
\mathbf{z}^{(0)} = (\mathbf{x},\, t) \in \mathcal{Z}, \quad \lambda_\text{HR}(t \mid \mathbf{x}) = \exp\!\left(\mathrm{KAN}_{\boldsymbol{\theta}}(\mathbf{x}, t)\right)
\label{eq:kaplan_hr}
\end{equation}

Observe that a single-layer \ac{KAPLAN-HR} model is structurally an additive B-spline \ac{PH} model, with a univariate spline for each covariate and for time: 
\begin{equation}
    \log \lambda_\text{HR}(t \mid \mathbf{x}) = \sum_{i=1}^{d} \varphi_i(x^{(i)}) + \varphi_t(t) \implies \lambda_\text{HR}(t \mid \mathbf{x}) = \underbrace{\exp\!\left(\varphi_t(t)\right)}_{\lambda_0(t)} \cdot \exp\!\left(\sum_{i=1}^{d} \varphi_i(x^{(i)})\right)
\end{equation}
For $D \geq 2$, the hidden representations combine time and covariate effects. For $D=2$:
\begin{equation}
    \log \lambda_\text{HR}(t \mid \mathbf{x}) = \sum_{j=1}^{d_1} \varphi_{2, j, 1}\!\left(\sum_{i=1}^{d} \varphi_{1, i, j}(x^{(i)}) + \varphi_{1, d_0, j}(t)\right)
\end{equation}
Network depth therefore removes both the additive and proportional-hazards restrictions, admitting covariate interactions and time-varying effects. The cumulative hazard $\Lambda_\text{HR}$ has no closed form for \ac{KAN}-modelled $\lambda_\text{HR}$, so we approximate it by numerical integration. We define a grid of time points $\tau_0 < \tau_1 < \cdots < \tau_K$ (with $\tau_0 \equiv 0$). Then, on each interval $[\tau_{k-1}, \tau_k]$ we apply a right-endpoint rectangle rule:
\begin{equation}
    \Delta\hat{\Lambda}[k \mid \mathbf{x}] = \lambda_\text{HR}(\tau_k \mid \mathbf{x}) \cdot (\tau_k - \tau_{k-1})
    \;\approx\; \int_{\tau_{k-1}}^{\tau_k} \lambda_\text{HR}(u \mid \mathbf{x})\,du
    \label{eq:hr_riemann}
\end{equation}
We use square bracket notation, such as $\hat{\Lambda}[k \mid \mathbf{x}]$, to distinguish discrete-time survival quantities. Accumulating these increments yields the discrete survival estimate:
\begin{equation}
    \hat{S}[k \mid \mathbf{x}] := \exp\!\left(-\hat{\Lambda}_\text{HR}[k \mid \mathbf{x}]\right) = \exp\!\left(-\sum_{m=1}^{k} \Delta\hat{\Lambda}[m \mid \mathbf{x}]\right)
    \label{eq:discrete_survival_from_hazard}
\end{equation}
The \ac{KAPLAN-HR} model is then trained by minimising the \ac{NLL} $\mathcal{L}_\text{HR}$ (\cref{eq:loss_hr}). We detail the implementation, configuration, and training procedure in \cref{app:implementation_details}.

\section{Theoretical Analysis}
\label{sec:theory}
We establish a rate of convergence of \ac{KAPLAN-HR} to the ground truth log-hazard function using the method of sieves \cite{shenConvergenceRateSieve1994} as evidence of our method's statistical correctness. Under the structural assumption that the true log-hazard admits a representation as a layered composition of smooth univariate functions, this convergence rate is independent of the input dimension, meaning \ac{KAPLAN-HR} mitigates the curse of dimensionality for true log-hazards in this problem class.

Denote by $\mathcal{X} := [0, 1]^d$ the covariate domain and $\mathcal{Z} := \mathcal{X} \times [0, 1] = [0, 1]^{d_0}$ the input domain including time. Further denote integer smoothness $r \ge 1$. We define $\mathcal{F}^{\mathrm{KAN}}_r$ the class of functions $g : \mathcal{Z} \to \mathbb{R}$ admitting a representation as an additive \ac{KAN} composition of univariate $W^{r, \infty}$ functions. Fix this composition's depth $D \ge 1$, input dimension $d_0$, and uniform hidden layer widths $d_1 = \cdots = d_{D-1} = Q$. Each edge's $W^{r, \infty}$-norm is at most $M$ and every composition $g \in \mathcal{F}^{\mathrm{KAN}}_r$ satisfies $\|g\|_{\infty, \mathcal{Z}} \le M_0 < M$ for some $M_0 \in [1, \infty)$.

Define the sieve $\mathcal{G}_n$ consisting of all \acp{KAN} with edges carrying B-splines of order $m > r$ on $J_n \asymp n^{1/(2r+1)}$ uniform interior knots. The splines' coefficients are bounded by $C_\theta$, such that every $g \in \mathcal{G}_n$ satisfies $\|g\|_{\infty, \mathcal{Z}} \leq M' := 2 M_0$.

We measure the distance between the true log-hazard $g^{*}$ and an estimate $g_n$ by the covariate-averaged Hellinger distance:
\begin{equation}
  d_H(\hat g_n, g^{*}) = \sqrt{\mathbb{E}_{X \sim P_X}\!\left[H^2(\mathbb{P}_{\hat g_n, X} \parallel \mathbb{P}_{g^{*}, X})\right]}
  \label{eq:hellinger}
\end{equation}
between conditional distributions $\mathbb{P}_{g, X}$ of $(Y, \Delta) \mid X$ under log-hazard $g$, where $P_X$ is the covariate distribution and $H^2(p, q) = \int (\sqrt{p} - \sqrt{q})^2$.

\paragraph{Assumptions}
\begin{itemize}
    \item[(C1)] $T \perp C \mid X$ independent censoring.
    \item[(C2)] Conditional on $X$, $T$ has (true) hazard $\lambda(t \mid \mathbf{x}) = e^{g^{*}(\mathbf{x}, t)}$ on $[0, 1]$ with $\|g^*\|_{\infty, \mathcal{Z}} \le M_0$ for some $M_0 \in [1, \infty)$.
    \item[(C3)] Conditional on $X$, $C \in [0, 1]$ has a continuous density $f_C(\cdot \mid \mathbf{x})$ on $[0, 1)$ together with a point mass $\alpha(\mathbf{x}) \in [0, 1]$ at $t = 1$, with $\int_0^1 f_C(s \mid \mathbf{x})\,ds + \alpha(\mathbf{x}) = 1$.
    \item[(C4)] The true log-hazard $g^{*}$ lies in $\mathcal{F}^{\mathrm{KAN}}_r(d_0, D, Q, M_0, M)$ with $M_0 < M$.
    \item[(C5)] The estimator $\hat g_n$ lies in $\mathcal{G}_n$ and falls short of the sieve log-likelihood supremum by at most a deterministic $\eta_n \le C_\eta\, n^{-2r/(2r+1)}\log n$ where $C_\eta \ge 0$ is an $n$-independent constant.
\end{itemize}  
Assumption (C2) positions the true log-hazard $g^{*}$ (the \ac{HR} parameterisation) and caps $\|g^{*}\|_{\infty, \mathcal{Z}}$. (C3) formalises the censoring mechanism and explicitly accommodates administrative censoring at $t=1$. (C4) is the structural assumption on $g^{*}$ that the sieve targets. (C5) puts $\hat g_n$ inside the sieve and limits the slack between $\hat g_n$ and the sieve optimum that the rate theorem can absorb.

\begin{theorem}[\ac{KAPLAN-HR} Hellinger rate]\label{thm:hr-rate}
Under (C1)-(C5): 
\begin{equation*}
      d_H\!\left(\hat g_n,\, g^{*}\right) = O_P\!\left(n^{-r/(2r+1)} \sqrt{\log n}\right)
\end{equation*}
\end{theorem}

Denote by $\|f\|_{L^2}$ the $L^2$ norm with $P_X$-weighted covariates and uniform time on $[0, \tau], \tau \in (0, 1)$: $\|f\|_{L^2}^2 := \mathbb{E}_{X \sim P_X} \int_0^\tau f(\mathbf{x}, t)^2 \, dt$. Further, assume censoring positivity $\inf_{\mathbf{x},\, t \in [0, \tau]} S_C(t \mid \mathbf{x}) \geq c, c > 0$, which is not required for \cref{thm:hr-rate}.

\begin{corollary}[$P_X$-weighted $L^2$ log-hazard rate]\label{cor:l2-hazard}
Under (C1)-(C5) and censoring positivity:
\begin{equation*}
    \|\hat g_n - g^{*}\|_{L^2} = O_P\!\left(n^{-r/(2r+1)} \sqrt{\log n}\right)
\end{equation*}
\end{corollary}
This follows from the curvature inequality $\|g - g^{*}\|_{L^2}^2 \lesssim d_H^2(g, g^{*})$ on $\|g\|_{\infty, \mathcal{Z}} \le M'$ under the additional censoring positivity hypothesis, applied to \cref{thm:hr-rate}.

\begin{corollary}[Survival function $L^2$ rate]\label{cor:ise-survival}
Under the hypotheses of \cref{cor:l2-hazard}, the survival estimator $\hat S_n(t \mid \mathbf{x}) = \exp\!\left(-\int_0^t e^{\hat g_n(\mathbf{x}, s)}\,ds\right)$ satisfies:
\begin{equation*}
  \|\hat S_n - S^{*}\|_{L^2} = O_P\!\left(n^{-r/(2r+1)} \sqrt{\log n}\right)
\end{equation*}
\end{corollary}
This follows from the Lipschitz stability of $g \mapsto \exp(-\int_0^\cdot e^{g(\mathbf{x}, s)}\,ds)$ on $\|g\|_{\infty, \mathcal{Z}} \le M'$ applied to \cref{cor:l2-hazard}.

\paragraph{Proof outline} \cref{thm:hr-rate} is proved by sieve maximum-likelihood estimation. The full proof is given in \cref{app:proofs}. We prove four propositions that supply the inputs that the rate theorem \cite[Thm. 3.4.12]{vandervaartWeakConvergenceEmpirical2023} requires for an $O_P$ rate in Hellinger distance (\cref{app:proof_propositions}). 
\begin{enumerate}
    \item \cref{prop:approximation}: Approximation error bound. Every target $g^{*} \in \mathcal{F}^{\mathrm{KAN}}_r$ has a sieve approximant $\bar g_n \in \mathcal{G}_n$ within $O(J_n^{-r})$. 
    \item \cref{prop:entropy}: Bracketing entropy bound. We quantify the complexity of the sieve $\mathcal{G}_n$ by upper-bounding its bracketing entropy at scale $\varepsilon$ by $O(p_n \log(p_n/\varepsilon))$, where $p_n$ is the sieve's parameter count.
    \item \cref{prop:bounded_likelihood_ratio}: Likelihood ratio bound. We transfer the proximity of $g^{*}$ and $\bar g_n$ to a bounded $\|p_0 / p_{\bar g_n}\|_\infty$.
    \item \cref{prop:bracketing_integral_bound}: Bracketing integral bound. We show the Hellinger bracketing integral is dominated by an explicit envelope $\widetilde J_n(\delta) \asymp \sqrt{J_n} \delta \sqrt{\log(J_n/\delta)}$, matching the rate theorem's shape requirements.
\end{enumerate}
Balancing these four bounds at $J_n \asymp n^{1/(2r+1)}$ yields the rate of \cref{thm:hr-rate} for the exact sieve maximiser. The estimator $\hat g_n$ is not an exact maximiser but an $\eta_n$-near-maximiser per assumption (C5). To accommodate this, we re-trace the derivation of the rate theorem through \cite[Thm. 3.4.1]{vandervaartWeakConvergenceEmpirical2023}. The parent theorem's slack-tolerance hypothesis absorbs $\eta_n$, transferring the rate to $\hat g_n$.

\paragraph{Pure B-splines vs SiLU residual} The class $\mathcal{F}^{\mathrm{KAN}}_r$ and sieve $\mathcal{G}_n$ define the \ac{KAN} edges as pure B-splines. However, the implementation includes a SiLU residual term in each edge (see \cref{fig:abstract}). Setting the residual term's trainable weight to zero embeds the theory's sieve $\mathcal{G}_n$ in the implementation's sieve, so the approximation rate of \cref{prop:approximation} transfers. The SiLU scalars are $O(1)$ extra parameters per edge contributing $O(\log n)$ to the bracketing entropy of \cref{prop:entropy} at scale $\varepsilon \asymp \delta_n$, dominated by the $O(J_n \log n)$ B-spline contribution. This dominance transfers to the bracketing integral envelope of \cref{prop:bracketing_integral_bound}. As a result, the rate exponent of \cref{thm:hr-rate} and of its corollaries is unaffected. 

\section{Related Work}
\label{sec:related_work}
\paragraph{Deep survival analysis} Deep learning has been widely applied to survival analysis \cite{wiegrebeDeepLearningSurvival2024, chenIntroductionDeepSurvival2024}, prominently by substituting \acp{MLP} into classical parameterisations \cite{katzmanDeepSurvPersonalizedTreatment2018, kvammeTimetoeventPredictionNeural2019, leeDeepHitDeepLearning2018}, but also with neural ODEs \cite{tangSODENScalableContinuoustime2022} and transformers \cite{wangSurvTRACETransformersSurvival2022}. Influential models include DeepSurv \cite{katzmanDeepSurvPersonalizedTreatment2018}, DeepHit \cite{leeDeepHitDeepLearning2018}, and \ac{DSM} \cite{nagpalAutonsurvivalOpenSourcePackage}. We refer readers to \citet{chenIntroductionDeepSurvival2024} and \citet{wiegrebeDeepLearningSurvival2024} for comprehensive reviews. Recent work has also applied \acp{KAN} to survival analysis \cite{knottenbeltCoxKANKolmogorovArnoldNetworks2025, jose2025kanaftinterpretablenonlinearsurvival, cheng2025extendingcoxproportionalhazards, mastroleo2026survkanfullyparametricsurvival}. A subset of deep learning survival analysis studies provides theoretical analyses of their proposed estimators' convergence and asymptotic normality. These include Deep Extended Hazards \cite{zhongDeepExtendedHazard2021}, Neural Frailty Machines \cite{wuNeuralFrailtyMachine2023}, and Deep Partially Linear Transformations \cite{yinDeepPartiallyLinear2025}. 

\paragraph{Kolmogorov-Arnold Network approximation guarantees} \citet{liuKANKolmogorovarnoldNetworks2025}'s initial \ac{KAN} paper establishes a dimension-free approximation upper bound for B-spline \acp{KAN} when the target admits a representation as an additive composition of smooth univariate edges. \citet{liuRateConvergenceKolmogorovArnold2025} formalise this analysis for two-layer \acp{KAN}, deriving a matching lower bound and thus establishing minimax optimality. \citet{kratsios2025approximationratesbesovnorms} introduce an augmented \ac{KAN} architecture with residual connections between layers and show that it is an optimal approximator for Besov targets. Complementary lines of work in \ac{KAN} theory include comparisons and mutual representation between \acp{KAN} and \acp{MLP} \cite{shuklaComprehensiveFAIRComparison2024, wang2025on}, guarantees on gradient descent optimisation \cite{gaoConvergenceStochasticGradient2025}, and generalisation bounds \cite{zhang2025generalization,li2025generalization}.

\section{Experiments}
\label{sec:experiments}
\subsection{Simulations}
\label{sec:simulations}
\begin{figure}[tbp]
    \centering
    \includegraphics[width=\linewidth]{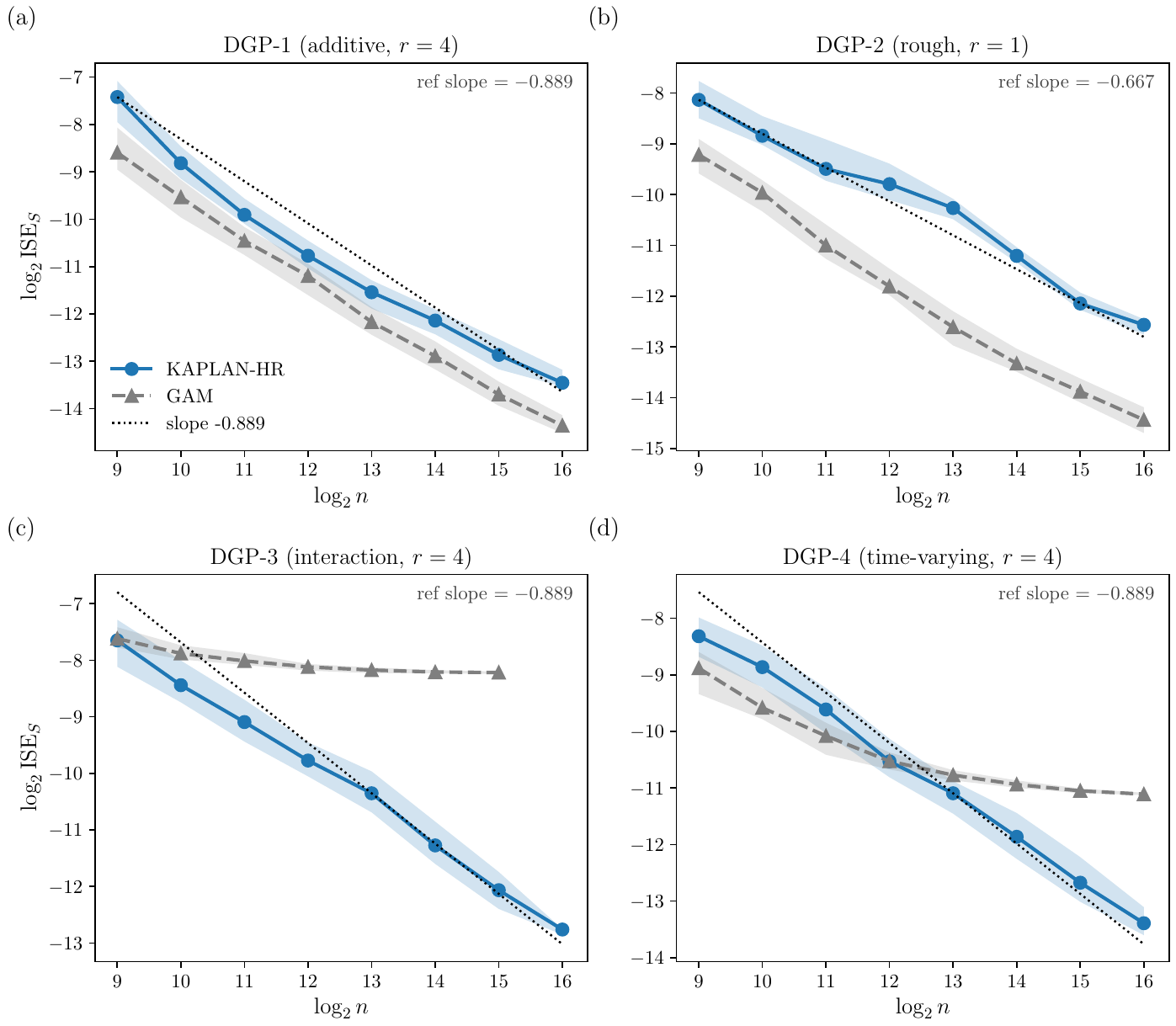}
    \caption{Log-log convergence of $\mathrm{ISE}_S$ vs. $n$ for \ac{KAPLAN-HR} and \ac{GAM}. Median and \ac{IQR} over $R=100$ repetitions. The dashed line marks the predicted convergence slope $-2r/(2r+1)$.}
    \label{fig:sim_convergence}
\end{figure}

We conduct simulation experiments to test the finite-sample validity of the convergence rates established in \cref{sec:theory}. We define four \acp{DGP} with different log-hazards: \ac{DGP}-1 is additive in $(x, t)$ and thus representable by a single \ac{KAN} layer. \ac{DGP}-2 is also additive but includes the tent term $|2x_1 - 1|$, which alters the smoothness exponent $r$. \ac{DGP}-3 contains an interaction term $\sin(\pi(x_1 + 0.5x_2))$, which an additive model cannot represent exactly, thus requiring at least two \ac{KAN} layers. \ac{DGP}-4 contains a time interaction term $x_1 \sin(\pi t)$, also requiring multiple layers. The \ac{DGP} formulae and experimental details are given in \cref{app:experimental_details_simulation}. 

For each \ac{DGP} we generate $R = 100$ independent training sets per \ac{DGP} at sample sizes $n \in \{2^9, 2^{10}, \ldots, 2^{16}\}$, plus a shared test set. We draw i.i.d. covariates $x_1, \ldots, x_5 \sim \mathrm{Uniform}(0,1)$ ($x_3, x_4, x_5$ are intentionally left as noise) and generate continuous event times $T$ by inverting the cumulative hazard. Censoring times are drawn independent of $T$ at an overall censoring rate of $30\%$, including administrative censoring at a fixed horizon. 

For each data realisation, we measure the \ac{ISE} of the predicted survival function against the true survival: 
\begin{equation}
\mathrm{ISE}_S = \|\hat S - S^*\|_{L^2}^2
\end{equation}
\cref{cor:ise-survival} predicts that $\log \mathrm{ISE}_S$ decays with $\log n$ at slope $-2r/(2r+1)$ for smoothness exponent $r$ (note the numerator is $-2r$ rather than $-r$ due to squaring). As a finite-sample baseline, we fit and evaluate \acp{GAM} on the same data. 

\paragraph{Results}
\Cref{fig:sim_convergence} reports the median $\log_2 \mathrm{ISE}_S$ against $\log_2 n$ on the held-out test set for \ac{KAPLAN-HR} and \ac{GAM}. The reference lines mark the slope predicted by \cref{cor:ise-survival}. \ac{KAPLAN-HR} tracks the predicted rate on all four \acp{DGP}, across the full range of $n$ on \ac{DGP}-1 and \ac{DGP}-2, and at larger $n$ on \ac{DGP}-3 and \ac{DGP}-4. This provides empirical support for the rate analysis in \cref{sec:theory}. Additionally, we observe that \ac{GAM} converges on \ac{DGP}-1 and \ac{DGP}-2, where it is well-specified, but plateaus on \ac{DGP}-3 and \ac{DGP}-4, whereas \ac{KAPLAN-HR} continues to descend. This demonstrates the advantage of \ac{KAPLAN-HR}'s compositional architecture for capturing covariate and time interactions outside the additive class.

\subsection{Real-World Clinical Datasets}
\label{sec:benchmarks}

\begin{figure}[tbp]
    \centering
    \includegraphics[width=\linewidth]{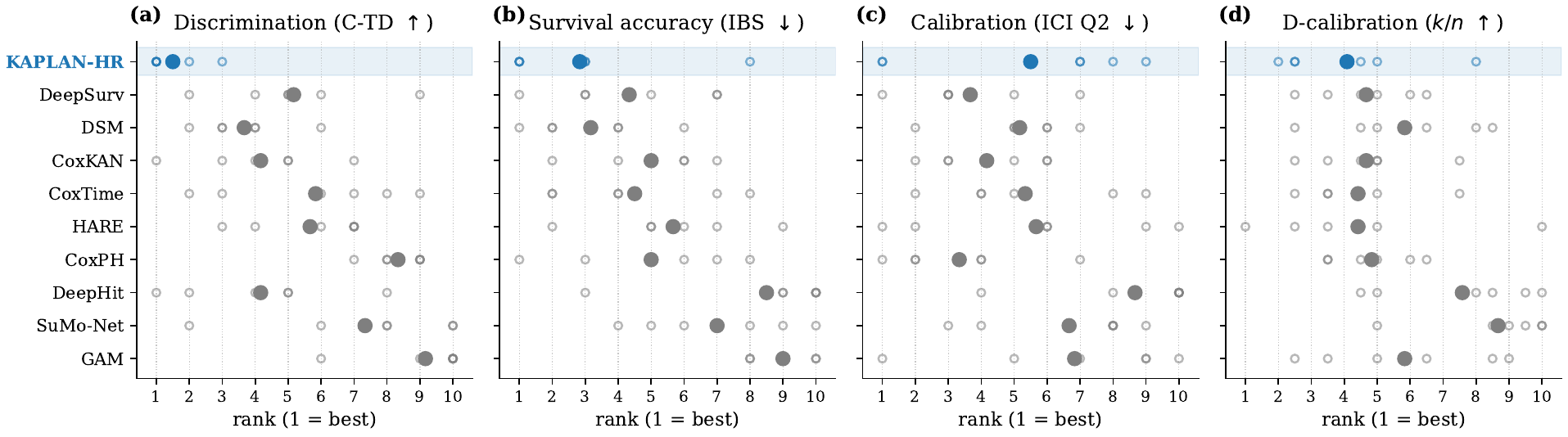}
    \caption{Ranking of \ac{KAPLAN-HR} vs. baselines across datasets for time-dependent concordance \ac{C-TD} ($\uparrow$), \ac{IBS} ($\downarrow$), \ac{ICI} at the median event time ($\downarrow$), and \ac{D-CAL} pass count  out of 10 ($\uparrow$). Small markers show per-dataset ranks, and large markers show the mean across datasets.} 
    \label{fig:empirical_results}
\end{figure}
We benchmark \ac{KAPLAN-HR} on six standard clinical survival analysis datasets. We evaluate predictive performance in terms of discrimination, via Antolini's \ac{C-TD} \cite{antoliniTimedependentDiscriminationIndex2005}; survival function accuracy, via the \ac{IBS} \cite{grafAssessmentComparisonPrognostic1999}; and calibration, via the median event time \ac{ICI} \cite{austinGraphicalCalibrationCurves2020} and the \ac{D-CAL} statistic \cite{haiderEffectiveWaysBuild2020}. Performance is compared against nine baselines spanning classical and deep learning approaches. 

\cref{app:experimental_details_benchmark} details our experimental protocol. Each dataset is split ten times into training, validation, and testing sets $64:16:20$. We report the mean and standard deviation across splits for \ac{C-TD}, \ac{IBS}, and \ac{ICI}, and the number of splits passing the \ac{D-CAL} test at $p > 0.05$. We tune \ac{KAPLAN-HR} and all tunable baseline models via Bayesian hyperparameter optimisation to maximise \ac{C-TD}. 

\paragraph{Results} 
\Cref{tab:headline-ctd-hr,tab:headline-ibs-hr,tab:headline-cal-hr} report \ac{KAPLAN-HR}'s scores against the highest-scoring baseline per dataset, and \cref{fig:empirical_results} shows \ac{KAPLAN-HR}'s mean rank on each metric compared to the other baseline models. \ac{KAPLAN-HR} attains the best \ac{C-TD} on 4 of 6 datasets, the best \ac{IBS} on 3 of 6 datasets, and the best \ac{ICI} on 2 of 6 datasets. We assess the statistical significance of each comparison via two-sided Wilcoxon signed-rank test on the paired per-split scores, with Holm-Bonferroni correction across the six datasets (column $p$ in each table). The \ac{C-TD} improvement on the FLCHAIN dataset is significant at $\alpha = 0.05$. None of the remaining comparisons reaches significance in either direction. This supports that \ac{KAPLAN-HR} is matched with the highest-scoring statistical or neural baseline per dataset in most comparisons, with one significant improvement and no significant degradations. \cref{app:benchmark_results} presents the complete benchmark results. 

\begin{table}[htbp]
\centering
\caption{Discrimination (C-TD, $\uparrow$) for KAPLAN-HR vs best-scoring baseline per dataset. $^{*}\!p<0.05$ (Wilcoxon signed-rank, Holm-corrected across datasets within metric).}
\label{tab:headline-ctd-hr}
\small
\begin{tabular}{l r ll r c}
\toprule
 & \textbf{KAPLAN-HR} & \multicolumn{2}{c}{\textbf{Best-Scoring Baseline}} & & \\
\cmidrule(lr){2-2} \cmidrule(lr){3-4}
\textbf{Dataset} & C-TD & Model & C-TD & $\Delta$ & $p$ \\
\midrule
  METABRIC & $0.6763 \pm 0.0161$ & CoxTime & $0.6729 \pm 0.0159$ & $+0.0033$ & 1.0000 \\
  MIMIC-III & $0.8114 \pm 0.0098$ & DeepHit & $0.8102 \pm 0.0084$ & $+0.0012$ & 1.0000 \\
  RotGBSG & $0.6671 \pm 0.0213$ & DSM & $0.6620 \pm 0.0226$ & $+0.0051$ & 1.0000 \\
  FLCHAIN & $0.7959 \pm 0.0128$ & DeepSurv & $0.7908 \pm 0.0128$ & $+0.0050$ & 0.0117$^{*}$ \\
  NWTCO & $0.7180 \pm 0.0243$ & CoxKAN & $0.7194 \pm 0.0257$ & $-0.0014$ & 1.0000 \\
  SUPPORT & $0.6706 \pm 0.0102$ & DeepHit & $0.6732 \pm 0.0127$ & $-0.0025$ & 0.9668 \\
\bottomrule
\end{tabular}
\end{table}

\begin{table}[htbp]
\centering
\caption{Survival function accuracy (IBS, $\downarrow$) for KAPLAN-HR vs best-scoring baseline per dataset. $^{*}\!p<0.05$ (Wilcoxon signed-rank, Holm-corrected across datasets within metric).}
\label{tab:headline-ibs-hr}
\small
\begin{tabular}{l r ll r c}
\toprule
 & \textbf{KAPLAN-HR} & \multicolumn{2}{c}{\textbf{Best-Scoring Baseline}} & & \\
\cmidrule(lr){2-2} \cmidrule(lr){3-4}
\textbf{Dataset} & IBS & Model & IBS & $\Delta$ & $p$ \\
\midrule
  METABRIC & $0.1658 \pm 0.0080$ & CoxPH & $0.1602 \pm 0.0069$ & $+0.0056$ & 0.4219 \\
  MIMIC-III & $0.1588 \pm 0.0169$ & DSM & $0.1841 \pm 0.0264$ & $-0.0253$ & 0.1641 \\
  RotGBSG & $0.1723 \pm 0.0102$ & CoxKAN & $0.1749 \pm 0.0108$ & $-0.0027$ & 0.5801 \\
  FLCHAIN & $0.0980 \pm 0.0027$ & DSM & $0.0981 \pm 0.0028$ & $-0.0001$ & 0.8633 \\
  NWTCO & $0.1021 \pm 0.0041$ & DSM & $0.0983 \pm 0.0036$ & $+0.0038$ & 0.1855 \\
  SUPPORT & $0.1834 \pm 0.0066$ & DeepSurv & $0.1828 \pm 0.0059$ & $+0.0007$ & 0.8633 \\
\bottomrule
\end{tabular}
\end{table}

\begin{table}[htbp]
\centering
\caption{Calibration (ICI ($\downarrow$) at median event time; D-cal = splits passing $p>0.05$) for KAPLAN-HR vs best-scoring baseline per dataset (by ICI). $^{*}\!p<0.05$ (Wilcoxon signed-rank, Holm-corrected across datasets within metric).}
\label{tab:headline-cal-hr}
\small
\begin{tabular}{l cc lcc r c}
\toprule
 & \multicolumn{2}{c}{\textbf{KAPLAN-HR}} & \multicolumn{3}{c}{\textbf{Best-Scoring Baseline}} & & \\\cmidrule(lr){2-3} \cmidrule(lr){4-6}
\textbf{Dataset} & ICI & D-cal & Model & ICI & D-cal & $\Delta$ ICI & $p$ \\
\midrule
  METABRIC & $0.0350 \pm 0.0114$ & 9/10 & DeepSurv & $0.0231 \pm 0.0074$ & 10/10 & $+0.0119$ & 0.0977 \\
  MIMIC-III & $0.0094 \pm 0.0017$ & 10/10 & HARE & $0.0079 \pm 0.0020$ & 10/10 & $+0.0015$ & 0.3926 \\
  RotGBSG & $0.0315 \pm 0.0087$ & 10/10 & CoxKAN & $0.0348 \pm 0.0163$ & 10/10 & $-0.0034$ & 0.8633 \\
  FLCHAIN & $0.0135 \pm 0.0053$ & 10/10 & DSM & $0.0145 \pm 0.0061$ & 10/10 & $-0.0009$ & 0.8633 \\
  NWTCO & $0.0362 \pm 0.0059$ & 10/10 & GAM & $0.0156 \pm 0.0153$ & 10/10 & $+0.0206$ & 0.1094 \\
  SUPPORT & $0.0433 \pm 0.0107$ & 7/10 & CoxPH & $0.0293 \pm 0.0081$ & 5/10 & $+0.0139$ & 0.0586 \\
\bottomrule
\end{tabular}
\end{table}

\section{Conclusion}
We introduced \ac{KAPLAN-HR}, a survival modelling approach that generalises established spline-based models with \acp{KAN}, able to capture covariate interactions and time-varying effects. Our theoretical analysis shows that, when the true log-hazard follows the \ac{KAN} structural assumption, the \ac{KAPLAN-HR} estimator attains a convergence rate that depends only on the smoothness of the representation and not on the covariate dimension up to a logarithmic factor, mitigating the curse of dimensionality. To our knowledge, this is the first formal convergence guarantee for a \ac{KAN}-based survival model. Simulation studies corroborate the predicted convergence rate and demonstrate that \ac{KAPLAN-HR} captures interactions that an additive \ac{GAM} cannot represent. Together with competitive performance on six clinical benchmarks, these results support that \acp{KAN} are well-suited for survival analysis.

\paragraph{Limitations} The convergence results we establish are informative about \ac{KAPLAN-HR}'s statistical properties but do not characterise finite-sample behaviour. Further, they rely on a structural assumption about the true log-hazard that is consistent with prior \ac{KAN} theory \cite{liuKANKolmogorovarnoldNetworks2025} but may not hold in all real-world situations. Finally, while our evaluation showed \ac{KAPLAN-HR} to be competitive with baselines, the standard benchmarking practice in survival analysis using medium-scale clinical datasets has known shortcomings \cite{wuNeuralFrailtyMachine2023}, with reported performance varying across publications. Larger-scale, standardised, and openly available benchmarks would benefit the survival analysis community.

\bibliographystyle{unsrtnat}
\bibliography{references}

@article{hastieGeneralizedAdditiveModels1995,
    title = {Generalized additive models for medical research},
    volume = {4},
    url = {https://doi.org/10.1177/096228029500400302},
    doi = {10.1177/096228029500400302},
    number = {3},
    journal = {Statistical Methods in Medical Research},
    author = {Hastie, Trevor and Tibshirani, Robert},
    year = {1995},
    pages = {187--196},
}

@article{kooperbergHazardRegression1995,
    title = {Hazard {Regression}},
    volume = {90},
    issn = {0162-1459},
    url = {https://doi.org/10.1080/01621459.1995.10476491},
    doi = {10.1080/01621459.1995.10476491},
    number = {429},
    journal = {Journal of the American Statistical Association},
    publisher = {Taylor \& Francis},
    author = {Kooperberg, Charles and Stone, Charles J. and Truong, Young K.},
    month = mar,
    year = {1995},
    pages = {78--94},
}

@article{woodGeneralizedAdditiveModels2025,
    title = {Generalized additive models},
    volume = {12},
    issn = {2326-831X},
    url = {https://www.annualreviews.org/content/journals/10.1146/annurev-statistics-112723-034249},
    doi = {https://doi.org/10.1146/annurev-statistics-112723-034249},
    number = {Volume 12, 2025},
    journal = {Annual Review of Statistics and Its Application},
    publisher = {Annual Reviews},
    author = {Wood, Simon N.},
    year = {2025},
    pages = {497--526},
}

@inproceedings{liuKANKolmogorovarnoldNetworks2025,
    title = {{KAN}: {Kolmogorov}-arnold networks},
    url = {https://openreview.net/forum?id=Ozo7qJ5vZi},
    booktitle = {Proceedings of the 2025 {International} {Conference} on {Learning} {Representations}},
    author = {Liu, Ziming and Wang, Yixuan and Vaidya, Sachin and Ruehle, Fabian and Halverson, James and Soljačić, Marin and Hou, Thomas Y. and Tegmark, Max},
    year = {2025},
}

@article{knottenbeltCoxKANKolmogorovArnoldNetworks2025,
    title = {{CoxKAN}: {Kolmogorov}-{Arnold} networks for interpretable, high-performance survival analysis},
    volume = {41},
    issn = {1367-4811},
    url = {https://doi.org/10.1093/bioinformatics/btaf413},
    doi = {10.1093/bioinformatics/btaf413},
    number = {8},
    urldate = {2025-12-15},
    journal = {Bioinformatics},
    author = {Knottenbelt, William and McGough, William and Wray, Rebecca and Zhang, Woody Zhidong and Liu, Jiashuai and Machado, Ines Prata and Gao, Zeyu and Crispin-Ortuzar, Mireia},
    month = aug,
    year = {2025},
    pages = {btaf413},
}

@misc{cheng2025extendingcoxproportionalhazards,
    title = {Extending cox proportional hazards model with symbolic non-linear log-risk functions for survival analysis},
    url = {https://doi.org/10.48550/arXiv.2504.04353},
    publisher = {arXiv},
    author = {Cheng, Jiaxiang and Hu, Guoqiang},
    year = {2025},
    note = {arXiv: 2504.04353 [cs.LG]},
}

@misc{jose2025kanaftinterpretablenonlinearsurvival,
    title = {{KAN}-{AFT}: {An} interpretable nonlinear survival model integrating kolmogorov-arnold networks with accelerated failure time analysis},
    url = {https://doi.org/10.48550/arXiv.2512.20305},
    publisher = {arXiv},
    author = {Jose, Mebin and Francis, Jisha and Kattumannil, Sudheesh Kumar},
    year = {2025},
}

@misc{mastroleo2026survkanfullyparametricsurvival,
    title = {{SurvKAN}: a fully parametric survival model based on kolmogorov-arnold networks},
    url = {https://doi.org/10.48550/arXiv.2602.02179},
    doi = {10.48550/arXiv.2602.02179},
    publisher = {arXiv},
    author = {Mastroleo, Marina and Archetti, Alberto and Mastroleo, Federico and Matteucci, Matteo},
    year = {2026},
}

@article{katzmanDeepSurvPersonalizedTreatment2018,
    title = {{DeepSurv}: personalized treatment recommender system using a {Cox} proportional hazards deep neural network},
    volume = {18},
    issn = {1471-2288},
    url = {https://doi.org/10.1186/s12874-018-0482-1},
    doi = {10.1186/s12874-018-0482-1},
    number = {1},
    journal = {BMC Medical Research Methodology},
    author = {Katzman, Jared L. and Shaham, Uri and Cloninger, Alexander and Bates, Jonathan and Jiang, Tingting and Kluger, Yuval},
    month = feb,
    year = {2018},
    pages = {24},
}

@article{kvammeTimetoeventPredictionNeural2019,
    title = {Time-to-event prediction with neural networks and cox regression},
    volume = {20},
    url = {http://jmlr.org/papers/v20/18-424.html},
    doi = {10.48550/arXiv.1907.00825},
    number = {129},
    journal = {Journal of Machine Learning Research},
    author = {Kvamme, Håvard and Borgan, Ørnulf and Scheel, Ida},
    year = {2019},
    pages = {1--30},
}

@article{leeDeepHitDeepLearning2018,
    title = {{DeepHit}: {A} {Deep} {Learning} {Approach} to {Survival} {Analysis} {With} {Competing} {Risks}},
    volume = {32},
    url = {https://doi.org/10.1609/aaai.v32i1.11842},
    doi = {10.1609/aaai.v32i1.11842},
    number = {1},
    journal = {Proceedings of the AAAI Conference on Artificial Intelligence},
    author = {Lee, Changhee and Zame, William and Yoon, Jinsung and van der Schaar, Mihaela},
    year = {2018},
}

@article{tangSODENScalableContinuoustime2022,
    title = {{SODEN}: a scalable continuous-time survival model through ordinary differential equation networks},
    volume = {23},
    url = {http://jmlr.org/papers/v23/20-900.html},
    doi = {10.48550/arXiv.2008.08637},
    number = {34},
    journal = {Journal of Machine Learning Research},
    author = {Tang, Weijing and Ma, Jiaqi and Mei, Qiaozhu and Zhu, Ji},
    year = {2022},
    pages = {1--29},
}

@inproceedings{wangSurvTRACETransformersSurvival2022,
    address = {Northbrook, Illinois},
    series = {Bcb '22},
    title = {{SurvTRACE}: transformers for survival analysis with competing events},
    isbn = {978-1-4503-9386-7},
    url = {https://doi.org/10.1145/3535508.3545521},
    doi = {10.1145/3535508.3545521},
    booktitle = {Proceedings of the 13th {ACM} international conference on bioinformatics, computational biology and health informatics},
    publisher = {Association for Computing Machinery},
    author = {Wang, Zifeng and Sun, Jimeng},
    year = {2022},
}

@inproceedings{zhongDeepExtendedHazard2021,
    address = {Red Hook, NY, USA},
    series = {Nips '21},
    title = {Deep extended hazard models for survival analysis},
    isbn = {978-1-7138-4539-3},
    url = {https://doi.org/10.5555/3540261.3541419},
    doi = {10.5555/3540261.3541419},
    booktitle = {Proceedings of the 35th international conference on neural information processing systems},
    publisher = {Curran Associates Inc.},
    author = {Zhong, Qixian and Mueller, Jonas and Wang, Jane-Ling},
    year = {2021},
}

@inproceedings{wuNeuralFrailtyMachine2023,
    title = {Neural {Frailty} {Machine}: {Beyond} proportional hazard assumption in neural survival regressions},
    url = {https://openreview.net/forum?id=3Fc9gnR0fa},
    booktitle = {Thirty-seventh conference on neural information processing systems},
    author = {Wu, Ruofan and Qiao, Jiawei and Wu, Mingzhe and Yu, Wen and Zheng, Ming and LIU, Tengfei and Zhang, Tianyi and Wang, Weiqiang},
    year = {2023},
}

@article{yinDeepPartiallyLinear2025,
    title = {Deep partially linear transformation model for right-censored survival data},
    volume = {81},
    issn = {0006-341X},
    url = {https://doi.org/10.1093/biomtc/ujaf126},
    doi = {10.1093/biomtc/ujaf126},
    number = {4},
    urldate = {2026-04-24},
    journal = {Biometrics},
    author = {Yin, Junkai and Zhang, Yue and Yu, Zhangsheng},
    month = dec,
    year = {2025},
    pages = {ujaf126},
}

@article{wiegrebeDeepLearningSurvival2024,
    title = {Deep learning for survival analysis: a review},
    volume = {57},
    issn = {1573-7462},
    url = {https://doi.org/10.1007/s10462-023-10681-3},
    doi = {10.1007/s10462-023-10681-3},
    number = {3},
    journal = {Artificial Intelligence Review},
    author = {Wiegrebe, Simon and Kopper, Philipp and Sonabend, Raphael and Bischl, Bernd and Bender, Andreas},
    month = feb,
    year = {2024},
    pages = {65},
}

@misc{chenIntroductionDeepSurvival2024,
    title = {An {Introduction} to {Deep} {Survival} {Analysis} {Models} for {Predicting} {Time}-to-{Event} {Outcomes}},
    url = {https://doi.org/10.48550/arXiv.2410.01086},
    doi = {10.48550/arXiv.2410.01086},
    publisher = {arXiv},
    author = {Chen, George H},
    year = {2024},
}

@article{coxRegressionModelsLifeTables1972,
    title = {Regression {Models} and {Life}-{Tables}},
    volume = {34},
    issn = {00359246},
    url = {http://www.jstor.org/stable/2985181},
    number = {2},
    urldate = {2026-03-12},
    journal = {Journal of the Royal Statistical Society. Series B (Methodological)},
    publisher = {[Royal Statistical Society, Oxford University Press]},
    author = {Cox, D. R.},
    year = {1972},
    pages = {187--220},
}

@misc{noorizadegan2026practitionersguidekolmogorovarnoldnetworks,
    title = {A {Practitioner}'s {Guide} to {Kolmogorov}-{Arnold} {Networks}},
    url = {https://doi.org/10.48550/arXiv.2510.25781},
    doi = {10.48550/arXiv.2510.25781},
    publisher = {arXiv},
    author = {Noorizadegan, Amir and Wang, Sifan and Ling, Leevan and Dominguez-Morales, Juan P.},
    year = {2026},
}

@book{schumakerSplineFunctionsBasic2007,
    address = {Cambridge},
    edition = {3rd ed},
    series = {Cambridge {Mathematical} {Library}},
    title = {Spline {Functions}: {Basic} {Theory}},
    isbn = {978-0-521-70512-7},
    url = {https://research.ebsco.com/linkprocessor/plink?id=7b136a48-6a99-3fe0-a0a8-2b0fc23984fc},
    language = {eng},
    publisher = {Cambridge University Press},
    author = {Schumaker, Larry},
    month = jan,
    year = {2007},
}

@misc{kratsios2025approximationratesbesovnorms,
      title={Approximation Rates in Besov Norms and Sample-Complexity of Kolmogorov-Arnold Networks with Residual Connections}, 
      author={Kratsios, Anastasis and Furuya, Takashi},
      year={2025},
      url = {https://doi.org/10.48550/arXiv.2504.15110},
      doi = {10.48550/arXiv.2504.15110},
      publisher = {arXiv},
}

@article{antoliniTimedependentDiscriminationIndex2005,
    title = {A time-dependent discrimination index for survival data},
    volume = {24},
    issn = {0277-6715},
    url = {https://doi.org/10.1002/sim.2427},
    doi = {10.1002/sim.2427},
    number = {24},
    urldate = {2026-04-10},
    journal = {Statistics in Medicine},
    publisher = {John Wiley \& Sons, Ltd},
    author = {Antolini, Laura and Boracchi, Patrizia and Biganzoli, Elia},
    month = dec,
    year = {2005},
    pages = {3927--3944},
}

@article{grafAssessmentComparisonPrognostic1999,
    title = {Assessment and comparison of prognostic classification schemes for survival data},
    volume = {18},
    issn = {0277-6715},
    url = {https://doi.org/10.1002/(SICI)1097-0258(19990915/30)18:17/18%3C2529::AID-SIM274%3E3.0.CO;2-5},
    doi = {10.1002/(SICI)1097-0258(19990915/30)18:17/18%3C2529::AID-SIM274%3E3.0.CO;2-5},
    number = {17-18},
    journal = {Statistics in Medicine},
    author = {Graf, Erika and Schmoor, Claudia and Sauerbrei, Willi and Schumacher, Martin},
    year = {1999},
    pages = {2529--2545},
}

@article{austinGraphicalCalibrationCurves2020,
    title = {Graphical calibration curves and the integrated calibration index ({ICI}) for survival models},
    volume = {39},
    issn = {0277-6715},
    url = {https://doi.org/10.1002/sim.8570},
    doi = {10.1002/sim.8570},
    number = {21},
    urldate = {2026-01-26},
    journal = {Statistics in Medicine},
    publisher = {John Wiley \& Sons, Ltd},
    author = {Austin, Peter C. and Harrell Jr, Frank E. and van Klaveren, David},
    month = sep,
    year = {2020},
    pages = {2714--2742},
}

@article{haiderEffectiveWaysBuild2020,
    title = {Effective ways to build and evaluate individual survival distributions},
    volume = {21},
    issn = {1532-4435},
    number = {1},
    journal = {Journal of Machine Learning Research},
    publisher = {JMLR.org},
    author = {Haider, Humza and Hoehn, Bret and Davis, Sarah and Greiner, Russell},
    month = jan,
    year = {2020},
}

@article{foekensUrokinaseSystemPlasminogen2000,
    title = {The {Urokinase} {System} of {Plasminogen} {Activation} and {Prognosis} in 2780 {Breast} {Cancer} {Patients}},
    volume = {60},
    issn = {0008-5472},
    number = {3},
    urldate = {2026-10-04},
    journal = {Cancer Research},
    author = {Foekens, John A. and Peters, Harry A. and Look, Maxime P. and Portengen, Henk and Schmitt, Manfred and Kramer, Michael D. and Brünner, Nils and Jänicke, Fritz and Gelder, Marion E. Meijer-van and Henzen-Logmans, Sonja C. and van Putten, Wim L. J. and Klijn, Jan G. M.},
    month = feb,
    year = {2000},
    pages = {636--643},
}

@article{schumacherRandomized221994,
    title = {Randomized 2 x 2 trial evaluating hormonal treatment and the duration of chemotherapy in node-positive breast cancer patients. {German} {Breast} {Cancer} {Study} {Group}.},
    volume = {12},
    issn = {0732-183X},
    url = {https://doi.org/10.1200/JCO.1994.12.10.2086},
    doi = {10.1200/JCO.1994.12.10.2086},
    number = {10},
    urldate = {2026-04-10},
    journal = {Journal of Clinical Oncology},
    publisher = {Wolters Kluwer},
    author = {Schumacher, M and Bastert, G and Bojar, H and Hübner, K and Olschewski, M and Sauerbrei, W and Schmoor, C and Beyerle, C and Neumann, R L and Rauschecker, H F},
    month = oct,
    year = {1994},
    pages = {2086--2093},
}

@article{breslowDesignAnalysisTwoPhase1999,
    title = {Design and {Analysis} of {Two}-{Phase} {Studies} with {Binary} {Outcome} {Applied} to {Wilms} {Tumour} {Prognosis}},
    volume = {48},
    issn = {0035-9254},
    url = {https://doi.org/10.1111/1467-9876.00165},
    doi = {10.1111/1467-9876.00165},
    number = {4},
    urldate = {2026-10-04},
    journal = {Journal of the Royal Statistical Society Series C: Applied Statistics},
    author = {Breslow, N. E. and Chatterjee, N.},
    month = dec,
    year = {1999},
    pages = {457--468},
}

@article{dispenzieriUseNonclonalSerum2012,
    title = {Use of {Nonclonal} {Serum} {Immunoglobulin} {Free} {Light} {Chains} to {Predict} {Overall} {Survival} in the {General} {Population}},
    volume = {87},
    issn = {0025-6196},
    url = {https://doi.org/10.1016/j.mayocp.2012.03.009},
    doi = {10.1016/j.mayocp.2012.03.009},
    number = {6},
    urldate = {2026-04-10},
    journal = {Mayo Clinic Proceedings},
    publisher = {Elsevier},
    author = {Dispenzieri, Angela and Katzmann, Jerry A. and Kyle, Robert A. and Larson, Dirk R. and Therneau, Terry M. and Colby, Colin L. and Clark, Raynell J. and Mead, Graham P. and Kumar, Shaji and Melton, III, L. Joseph and Rajkumar, S. Vincent},
    month = jun,
    year = {2012},
    pages = {517--523},
}

@article{knausSUPPORTPrognosticModel1995,
    title = {The {SUPPORT} {Prognostic} {Model}: {Objective} {Estimates} of {Survival} for {Seriously} {Ill} {Hospitalized} {Adults}},
    volume = {122},
    issn = {0003-4819},
    url = {https://doi.org/10.7326/0003-4819-122-3-199502010-00007},
    doi = {10.7326/0003-4819-122-3-199502010-00007},
    number = {3},
    urldate = {2026-04-10},
    journal = {Annals of Internal Medicine},
    publisher = {American College of Physicians},
    author = {Knaus, William A. and Harrell, Frank E. and Lynn, Joanne and Goldman, Lee and Phillips, Russell S. and Connors, Alfred F. and Dawson, Neal V. and Fulkerson, William J. and Califf, Robert M. and Desbiens, Norman and Layde, Peter and Oye, Robert K. and Bellamy, Paul E. and Hakim, Rosemarie B. and Wagner, Douglas P.},
    month = feb,
    year = {1995},
    pages = {191--203},
}

@article{kvammeContinuousDiscretetimeSurvival2021,
    title = {Continuous and discrete-time survival prediction with neural networks},
    volume = {27},
    issn = {1572-9249},
    url = {https://doi.org/10.1007/s10985-021-09532-6},
    doi = {10.1007/s10985-021-09532-6},
    number = {4},
    journal = {Lifetime Data Analysis},
    author = {Kvamme, Håvard and Borgan, Ørnulf},
    month = oct,
    year = {2021},
    pages = {710--736},
}

@article{johnsonMIMICIIIClinicalDatabase2016,
    title = {{MIMIC}-{III} clinical database},
    url = {https://doi.org/10.13026/C2XW26},
    doi = {10.13026/C2XW26},
    journal = {PhysioNet},
    author = {Johnson, Alistair and Pollard, Tom and Mark, Roger},
    month = sep,
    year = {2016},
}

@article{purushothamBenchmarkingDeepLearning2018,
    title = {Benchmarking deep learning models on large healthcare datasets},
    volume = {83},
    issn = {1532-0464},
    url = {https://www.sciencedirect.com/science/article/pii/S1532046418300716},
    doi = {10.1016/j.jbi.2018.04.007},
    journal = {Journal of Biomedical Informatics},
    author = {Purushotham, Sanjay and Meng, Chuizheng and Che, Zhengping and Liu, Yan},
    month = jul,
    year = {2018},
    pages = {112--134},
}

@article{davidson-pilonLifelinesSurvivalAnalysis2019,
    title = {lifelines: survival analysis in {Python}},
    volume = {4},
    url = {https://doi.org/10.21105/joss.01317},
    doi = {10.21105/joss.01317},
    number = {40},
    journal = {Journal of Open Source Software},
    publisher = {The Open Journal},
    author = {Davidson-Pilon, Cameron},
    year = {2019},
    pages = {1317},
}

@article{woodSmoothingParameterModel2016,
    title = {Smoothing {Parameter} and {Model} {Selection} for {General} {Smooth} {Models}},
    volume = {111},
    issn = {0162-1459},
    url = {https://doi.org/10.1080/01621459.2016.1180986},
    doi = {10.1080/01621459.2016.1180986},
    number = {516},
    journal = {Journal of the American Statistical Association},
    publisher = {Taylor \& Francis},
    author = {Wood, Simon N. and Pya, Natalya and Säfken, Benjamin},
    month = oct,
    year = {2016},
    pages = {1548--1563},
}

@misc{nagpalAutonsurvivalOpenSourcePackage,
    title = {auton-survival: an {Open}-{Source} {Package} for {Regression}, {Counterfactual} {Estimation}, {Evaluation} and {Phenotyping} with {Censored} {Time}-to-{Event} {Data}},
    url = {https://doi.org/10.48550/arXiv.2204.07276},
    doi = {10.48550/arXiv.2204.07276},
    publisher = {arXiv},
    author = {Nagpal, Chirag and Potosnak, Willa and Dubrawski, Artur},
    year = {2022},
}

@article{nagpalDeepSurvivalMachines2021,
    title = {Deep {Survival} {Machines}: {Fully} {Parametric} {Survival} {Regression} and {Representation} {Learning} for {Censored} {Data} {With} {Competing} {Risks}},
    volume = {25},
    issn = {2168-2208},
    doi = {10.1109/JBHI.2021.3052441},
    number = {8},
    journal = {IEEE Journal of Biomedical and Health Informatics},
    author = {Nagpal, Chirag and Li, Xinyu and Dubrawski, Artur},
    month = aug,
    year = {2021},
    pages = {3163--3175},
}

@inproceedings{rindtSurvivalRegressionProper2022,
    series = {Proceedings of machine learning research},
    title = {Survival regression with proper scoring rules and monotonic neural networks},
    volume = {151},
    url = {https://proceedings.mlr.press/v151/rindt22a.html},
    doi = {10.48550/arXiv.2103.14755},
    booktitle = {Proceedings of the 25th international conference on artificial intelligence and statistics},
    publisher = {PMLR},
    author = {Rindt, David and Hu, Robert and Steinsaltz, David and Sejdinovic, Dino},
    editor = {Camps-Valls, Gustau and Ruiz, Francisco J. R. and Valera, Isabel},
    month = mar,
    year = {2022},
    pages = {1190--1205},
}

@misc{weightsandbiasesSweepsTuneHyperparameters2026,
    title = {Sweeps: {Tune} hyperparameters},
    url = {https://docs.wandb.ai/guides/sweeps/},
    urldate = {2026-04-10},
    publisher = {Weights and Biases},
    author = {{Weights and Biases}},
    year = {2026},
}

@article{polsterlScikitsurvivalLibraryTimetoevent2020,
    title = {scikit-survival: a library for time-to-event analysis built on top of scikit-learn},
    volume = {21},
    url = {http://jmlr.org/papers/v21/20-729.html},
    number = {212},
    journal = {Journal of Machine Learning Research},
    author = {Pölsterl, Sebastian},
    year = {2020},
    pages = {1--6},
}

@article{qiSurvivalEVALComprehensiveOpensource2024,
    title = {{SurvivalEVAL}: a comprehensive open-source python package for evaluating individual survival distributions},
    volume = {2},
    doi = {10.1609/aaaiss.v2i1.27713},
    journal = {Proceedings of the AAAI Symposium Series},
    author = {Qi, Shi-ang and Sun, Weijie and Greiner, Russell},
    month = jan,
    year = {2024},
    pages = {453--457},
}

@article{curtisGenomicTranscriptomicArchitecture2012a,
    title = {The genomic and transcriptomic architecture of 2,000 breast tumours reveals novel subgroups},
    volume = {486},
    issn = {1476-4687},
    url = {https://doi.org/10.1038/nature10983},
    doi = {10.1038/nature10983},
    number = {7403},
    journal = {Nature},
    author = {Curtis, Christina and Shah, Sohrab P. and Chin, Suet-Feung and Turashvili, Gulisa and Rueda, Oscar M. and Dunning, Mark J. and Speed, Doug and Lynch, Andy G. and Samarajiwa, Shamith and Yuan, Yinyin and Gräf, Stefan and Ha, Gavin and Haffari, Gholamreza and Bashashati, Ali and Russell, Roslin and McKinney, Steven and {METABRIC Group} and Langerød, Anita and Green, Andrew and Provenzano, Elena and Wishart, Gordon and Pinder, Sarah and Watson, Peter and Markowetz, Florian and Murphy, Leigh and Ellis, Ian and Purushotham, Arnie and Børresen-Dale, Anne-Lise and Brenton, James D. and Tavaré, Simon and Caldas, Carlos and Aparicio, Samuel},
    month = jun,
    year = {2012},
    pages = {346--352},
}

@misc{liuRateConvergenceKolmogorovArnold2025,
    title = {On the {Rate} of {Convergence} of {Kolmogorov}-{Arnold} {Network} {Regression} {Estimators}},
    url = {https://doi.org/10.48550/arXiv.2509.19830},
    doi = {10.48550/arXiv.2509.19830},
    publisher = {arXiv},
    author = {Liu, Wei and Chatzi, Eleni and Lai, Zhilu},
    year = {2025},
}

@book{vandervaartWeakConvergenceEmpirical2023,
    address = {New York},
    edition = {2},
    series = {Springer {Series} in {Statistics}},
    title = {Weak {Convergence} and {Empirical} {Processes}},
    isbn = {978-3-031-29038-1},
    url = {https://doi.org/10.1007/978-3-031-29040-4},
    doi = {10.1007/978-1-4757-2545-2},
    publisher = {Springer New York},
    author = {van der Vaart, Aad W. and Wellner, Jon A.},
    year = {2023},
}

@article{sudlowUKBiobankOpen2015,
    title = {{UK} {Biobank}: {An} {Open} {Access} {Resource} for {Identifying} the {Causes} of a {Wide} {Range} of {Complex} {Diseases} of {Middle} and {Old} {Age}},
    volume = {12},
    url = {https://doi.org/10.1371/journal.pmed.1001779},
    doi = {10.1371/journal.pmed.1001779},
    number = {3},
    journal = {PLOS Medicine},
    publisher = {Public Library of Science},
    author = {Sudlow, Cathie and Gallacher, John and Allen, Naomi and Beral, Valerie and Burton, Paul and Danesh, John and Downey, Paul and Elliott, Paul and Green, Jane and Landray, Martin and Liu, Bette and Matthews, Paul and Ong, Giok and Pell, Jill and Silman, Alan and Young, Alan and Sprosen, Tim and Peakman, Tim and Collins, Rory},
    month = mar,
    year = {2015},
    pages = {e1001779},
}

@book{aliprantisInfiniteDimensionalAnalysis1999,
    address = {Heidelberg},
    series = {Studies in {Economic} {Theory}},
    title = {Infinite {Dimensional} {Analysis}},
    url = {https://doi.org/10.1007/978-3-662-03961-8},
    doi = {10.1007/978-3-662-03961-8},
    publisher = {Springer Berlin},
    author = {Aliprantis, Charalambos D. and Border, Kim C.},
    year = {1999},
}

@article{shenConvergenceRateSieve1994,
    title = {Convergence {Rate} of {Sieve} {Estimates}},
    volume = {22},
    url = {https://doi.org/10.1214/aos/1176325486},
    doi = {10.1214/aos/1176325486},
    number = {2},
    journal = {The Annals of Statistics},
    author = {Shen, Xiaotong and Wong, Wing Hung},
    month = jun,
    year = {1994},
    pages = {580--615},
}

@article{shuklaComprehensiveFAIRComparison2024,
    title = {A comprehensive and {FAIR} comparison between {MLP} and {KAN} representations for differential equations and operator networks},
    volume = {431},
    issn = {0045-7825},
    url = {https://www.sciencedirect.com/science/article/pii/S0045782524005462},
    doi = {10.1016/j.cma.2024.117290},
    journal = {Computer Methods in Applied Mechanics and Engineering},
    author = {Shukla, Khemraj and Toscano, Juan Diego and Wang, Zhicheng and Zou, Zongren and Karniadakis, George Em},
    month = nov,
    year = {2024},
    pages = {117290},
}

@article{gaoConvergenceStochasticGradient2025,
    title = {On the convergence of (stochastic) gradient descent for kolmogorov–arnold networks},
    volume = {71},
    doi = {10.1109/TIT.2025.3588401},
    number = {9},
    journal = {IEEE Transactions on Information Theory},
    author = {Gao, Yihang and Tan, Vincent Y. F.},
    year = {2025},
    pages = {7270--7291},
}

@inproceedings{zhang2025generalization,
    title = {Generalization bounds and model complexity for kolmogorov–arnold networks},
    url = {https://openreview.net/forum?id=q5zMyAUhGx},
    booktitle = {The thirteenth international conference on learning representations},
    author = {Zhang, Xianyang and Zhou, Huijuan},
    year = {2025},
}

@inproceedings{li2025generalization,
    title = {Generalization bounds for {Kolmogorov}-{Arnold} {Networks} ({KANs}) and enhanced {KANs} with lower {Lipschitz} complexity},
    url = {https://openreview.net/forum?id=b387eWFV3V},
    booktitle = {The thirty-ninth annual conference on neural information processing systems},
    author = {Li, Pengqi and Ding, Lizhong and Fu, Jiarun and {ChunhuiZhang} and Wang, Guoren and Yuan, Ye},
    year = {2025},
}

@inproceedings{wang2025on,
    title = {On the expressiveness and spectral bias of {KANs}},
    url = {https://openreview.net/forum?id=ydlDRUuGm9},
    booktitle = {The thirteenth international conference on learning representations},
    author = {Wang, Yixuan and Siegel, Jonathan W. and Liu, Ziming and Hou, Thomas Y.},
    year = {2025},
}

\clearpage 

\appendix
\crefalias{section}{appendix}
\crefalias{subsection}{appendix}

\section{Proofs}
\label{app:proofs}
\subsection{Notation} We introduce some notation that features frequently in the proof: For a function class $\mathcal{F}$, denote the covering number $N(\varepsilon, \mathcal{F}, \|\cdot\|)$ and the bracketing number $N_{[\,]}(\varepsilon, \mathcal{F}, \|\cdot\|)$ at scale $\varepsilon$ with respect to norm $\|\cdot\|$. Denote also the bracketing integral:
\begin{equation}\label{eq:setup-Jbra}
  \mathcal{J}_{[\,]}(\delta, \mathcal{F}, \|\cdot\|) = \int_0^\delta \sqrt{1 + \log N_{[\,]}(\varepsilon, \mathcal{F}, \|\cdot\|)}\, d\varepsilon.
\end{equation}

We use the notation $a \lesssim b$ to mean $a \leq Cb$ for constant $C > 0$, and $a \asymp b$ to mean $c_1 b \leq a \leq c_2  b$. 

\subsection{Setup}\label{sec:appendix-setup}
As defined in \cref{sec:theory}, let the class $\mathcal{F}^{\mathrm{KAN}}_r$ with architecture parameters $(d_0, D, Q, M_0, M)$ with smoothness $r \in \mathbb{N}_{\ge 1}$, spline order $m > r$, bounding constants $M_0 \in [1, \infty)$, $M_0 < M$, and depth $D \geq 1$. Let also the sieve $\mathcal{G}_n$ with knot count $J_n \asymp n^{1/(2r+1)}$ and sup-norm cap $M' := 2 M_0$. We assume conditions (C1)-(C5) and the censoring law $f_C, \alpha, S_C$ from (C3). Finally, we use the covariate-averaged Hellinger metric $d_H$ from \cref{eq:hellinger}. 

$\mathcal{X} := [0, 1]^d$ is the covariate domain, and $\mathcal{Z} := \mathcal{X} \times [0, 1] = [0, 1]^{d_0}$ is the model input domain (covariates and time). Each \ac{KAN} layer $\Phi_\ell, \ell \in \{1, \ldots, D\}$ has domain $\mathcal{Z}_\ell$ with $\mathcal{Z}_1 = \mathcal{Z}$.

The sieve's total parameter count is $p_n = E(J_n + m)$ where $E := \sum_{\ell=1}^D d_{\ell - 1} d_\ell$ is the total edge count. Each sieve edge is an order-$m$ B-spline on $J_n$ uniform interior knots ($m$-fold boundary knots) parameterised by a coefficient vector in $\mathbb{R}^{J_n + m}$ with $\|\cdot\|_\infty \le C_\theta$. Denote the corresponding coefficient box $\Theta_n := [-C_\theta, C_\theta]^{p_n}$. 

Define the dominating measure on $\mathcal{Z}$:
\begin{equation*}
    \mu = P_X \otimes \mathrm{Leb}_{[0, 1]} \otimes (\delta_0 + \delta_1) \;+\; P_X \otimes \delta_1 \otimes \delta_0 
\end{equation*}
where the left term corresponds to events and censoring before $Y=1$, and the right term corresponds to administrative censoring. Under (C1) and (C3), the joint density $p_g$ of $(X, Y, \Delta)$ with respect to $\mu$ is
\begin{equation}
    \label{eq:setup-pg}
    \begin{cases}
        p_g(\mathbf{x}, y, 1) = S_C(y \mid \mathbf{x})\, e^{g(\mathbf{x}, y)}\, S_g(y \mid \mathbf{x}) & \Delta = 1,\, y \in [0, 1) \\
        p_g(\mathbf{x}, y, 0) = f_C(y \mid \mathbf{x})\, S_g(y \mid \mathbf{x}) & \Delta = 0,\, y \in [0, 1) \\
        p_g(\mathbf{x}, 1, 0) = \alpha(\mathbf{x})\, S_g(1 \mid \mathbf{x}) & \Delta = 0,\, y = 1 \\
        p_g(\mathbf{x}, 1, 1) := 0 & \Delta = 1,\, y = 1
    \end{cases}
\end{equation}
where $S_g$ denotes the survival function induced by log-hazard $g$. 

The empirical log-likelihood decomposes into the g-dependent empirical criterion $\mathcal{L}_n(g)$ and the g-invariant remainder $R_n$ which depends only on the censoring mechanism and is finite. 
\begin{equation}
    \label{eq:log-likelihood}
    \frac{1}{n}\sum_i \log p_g (\mathbf{x}_i, y_i, \Delta_i) = \mathcal{L}_n(g) + R_n 
\end{equation}

The joint Hellinger distance on densities w.r.t. $\mu$ is
\begin{equation}\label{eq:setup-H}
  H(p, q) = \left(\int (\sqrt p - \sqrt q)^2\, d\mu\right)^{1/2}, \qquad H^2(p, q) = \int (\sqrt p - \sqrt q)^2\, d\mu.
\end{equation}
The covariate-averaged Hellinger $d_H$ from \cref{eq:hellinger} relates to $H$ via $\mu = P_X \otimes \nu$ (with $\nu = \mathrm{Leb}_{[0, 1]} \otimes (\delta_0 + \delta_1) + \delta_1 \otimes \delta_0$): both densities $p_g, p_{g'}$ have covariate marginal $P_X$, so:
\begin{equation}\label{eq:setup-dH}
  d_H^2(g, g') = \mathbb{E}_{X \sim P_X}\left[H^2(\mathbb{P}_{g, X} \parallel \mathbb{P}_{g', X})\right] = \int (\sqrt{p_g} - \sqrt{p_{g'}})^2\, d\mu = H^2(p_g, p_{g'}).
\end{equation}
Hence $H(p_g, p_{g'}) = d_H(g, g')$ throughout the proof.

\subsection{Propositions}
\label{app:proof_propositions}
The following propositions are used in the main theorem proof. The proposition proofs are given in \cref{app:proposition_proofs}.
\begin{proposition}[Approximation bound]
    \label{prop:approximation}
    There exist constants $J_0, C_{\mathrm{app}} \in (0, \infty)$ independent of $n$ such that, for every $g^* \in \mathcal{F}^{\mathrm{KAN}}_r$ and every $J_n \ge J_0$, there exists a deterministic sieve approximant $\bar{g}_n \in \mathcal{G}_n$ with:
    \begin{equation}
      \|\bar g_n - g^*\|_{\infty, \mathcal{Z}} \leq C_{\mathrm{app}}\, J_n^{-r}.
    \end{equation}
\end{proposition}

\begin{proposition}[Bracketing entropy bound]
    \label{prop:entropy}
    There exist constants $p_{\min}, C_{\mathrm{ent}} \in (0, \infty)$ independent of $n$ such that for every $\varepsilon \in (0, 2M']$ and every $n$ for which $p_n \geq p_{\min}$, the bracketing entropy:
\begin{equation}
    \log N_{[\,]}(\varepsilon, \mathcal{G}_n, \|\cdot\|_{\infty, \mathcal{Z}}) \leq C_{\mathrm{ent}} p_n \log(p_n/\varepsilon)
\end{equation}
\end{proposition}

\begin{proposition}[Likelihood ratio bound]
    \label{prop:bounded_likelihood_ratio} 
    Let constant $B \in [0, \infty)$. For every pair of measurable log-hazards $g_1, g_2 : \mathcal{Z} \rightarrow \mathbb{R}$ with $\|g_i\|_{\infty, \mathcal{Z}} \leq B$ ($i = 1,2$):
    \begin{equation}
        \left|\log \frac{p_{g_1}(\mathbf{x}, y, \Delta)}{p_{g_2}(\mathbf{x}, y, \Delta)}\right| \leq (1 + e^B) \|g_1 - g_2\|_{\infty, \mathcal{Z}}
    \end{equation}
    for every $(\mathbf{x}, y, \Delta)$ in the $\mu$ support.
\end{proposition}

\begin{proposition}[Bracketing integral bound]
    \label{prop:bracketing_integral_bound}
    There exist constants $K, J^* \in (0, \infty)$ independent of $n$ such that for every $n$ with $J_n \ge J^*$, the function:
    \begin{equation}
        \widetilde J_n(\delta) := K \sqrt{J_n} \delta\sqrt{\log(J_n/\delta) + 1}, \qquad \delta \in (0, \sqrt 2]
    \end{equation} satisfies: 
    \begin{enumerate}[label=(\roman*)]
        \item $\widetilde J_n$ is strictly increasing
        \item $\widetilde J_n(\delta) / \delta^{3/2}$ is strictly decreasing
        \item $\mathcal{J}_{[\,]}(\delta, \mathcal{P}_n, H) \leq \widetilde J_n(\delta)$
    \end{enumerate}
\end{proposition}

\subsection{Proof of Theorem 1} Let the sieve density class $\mathcal{P}_n := \{p_g : g \in \mathcal{G}_n\}$. For notational ease denote $\hat p_n = p_{\hat g_n}$, and $p_0 = p_{g^{*}}$. \cref{thm:hr-rate} is proven in two parts. The first part is invoking \cite[Thm. 3.4.12]{vandervaartWeakConvergenceEmpirical2023}: For the exact sieve \ac{MLE} over $\mathcal{P}_n$, $H(\cdot, p_0) = O_P(\delta_n)$ conditional on: 
\begin{enumerate}[label=(\roman*)]
    \item \emph{Bounded ratio.} A sieve approximant density $p_n^{\circ} \in \mathcal{P}_n$ and constant $M_{\mathrm{ratio}} \in [1, \infty)$ such that $\|p_0/p_n^{\circ}\|_\infty \le M_{\mathrm{ratio}}$.
    \item \emph{Bounded bracketing integral} (\cite[Eq. 3.4.11]{vandervaartWeakConvergenceEmpirical2023} clause 1). $\mathcal{J}_{[\,]}(\delta_n, \mathcal{P}_n, H) \lesssim \sqrt n\, \delta_n^2$ at some $\delta_n \in (0, \sqrt 2\,]$.
    \item \emph{Discrepancy threshold} (\cite[Eq. 3.4.11]{vandervaartWeakConvergenceEmpirical2023} clause 2). $\delta_n \gtrsim H(p_n^{\circ}, p_0)$.
\end{enumerate}
Per assumption (C5), however, $\hat p_n$ is not the exact \ac{MLE} but rather an $\eta_n$-near-MLE. Therefore second part of the proof is to extend Thm. 3.4.12 to accommodate the near-maximiser slack by carrying it through the theorem's original derivation from \cite[Thm. 3.4.1]{vandervaartWeakConvergenceEmpirical2023} and \cite[Lem. 3.4.10]{vandervaartWeakConvergenceEmpirical2023}. 

Let measurable selector $\hat g_n \in \mathcal{G}_n$ such that $H(\hat p_n, p_0)$ is Borel measurable (given by \cref{lem:measurable_selector}). This lets us collapse Thm. 3.4.1's $O_P^*$ bound to $O_P$. Constrain $J_n \geq \max(J_0, J^{*}) =: J^{*}_{\mathrm{rate}}$ where $J_0$ is the minimum $J_n$ from \cref{prop:approximation} and $J^{*}$ is the same from \cref{prop:bracketing_integral_bound}. 

\paragraph{Step 1 - Sieve approximant} Given target $g^* \in \mathcal{F}^\mathrm{KAN}_r$ define the sieve approximant $\bar{g}_n \in \mathcal{G}_n$ such that $\|\bar{g}_n - g^{*}\|_{\infty, \mathcal{Z}} \leq C_{\mathrm{app}}J_n^{-r}$ via \cref{prop:approximation}. Denote the sieve approximant density $p_n^{\circ} = p_{\bar{g}_n} \in \mathcal{P}_n$. Both $g^{*}$ and $\bar g_n$ are sup-norm bounded by $M'$ from their definition. Invoke \cref{lem:sup_norm_to_hellinger} at $B = M'$:
\begin{equation}
    \label{eq:rate-approx}
    H^2(p_n^{\circ}, p_0) \leq C_h(M')\, \|\bar g_n - g^{*}\|_{\infty, \mathcal{Z}}^2 \leq C_h(M')\, C_{\mathrm{app}}^2\, J_n^{-2r}
\end{equation}
Where $C_h(\cdot)$ is a constant defined in \cref{lem:sup_norm_to_hellinger}. Setting $\delta_n^{(a)} = \sqrt{C_h(M')}\, C_{\mathrm{app}}\, J_n^{-r}$ gives $H(p_n^{\circ}, p_0) \le \delta_n^{(a)}$.

\paragraph{Step 2 - Bounded ratio} \cref{prop:bounded_likelihood_ratio} at $B = M'$ gives:
\begin{equation*}
    \left|\log \frac{p_0}{p_n^{\circ}}\right| \leq (1 + e^{M'}) \|\bar g_n - g^{*}\|_{\infty, \mathcal{Z}} \leq (1 + e^{M'})\, C_{\mathrm{app}}\, J_n^{-r}
\end{equation*}
on the $\mu$ support. Exponentiate and monotonise via $J_n \geq J^{*}_{\mathrm{rate}}$:
\begin{equation}
    \label{eq:rate-ratio}
    \|p_0 / p_n^{\circ}\|_\infty \leq M_{\mathrm{ratio}} := e^{\alpha_0}, \qquad \alpha_0 = (1 + e^{M'})\, C_{\mathrm{app}}\, (J^{*}_{\mathrm{rate}})^{-r}
\end{equation}

\paragraph{Step 3 - Bracketing integral} For $\delta \in (0, \sqrt 2\,]$, apply \cref{prop:bracketing_integral_bound}:
\begin{equation}
    \label{eq:rate-bracket}
    \mathcal{J}_{[\,]}(\delta, \mathcal{P}_n, H) \leq \widetilde J_n(\delta) := K \sqrt{J_n} \delta \sqrt{\log(J_n / \delta) + 1},
\end{equation}
with $\widetilde J_n$ strictly increasing on $(0, \sqrt 2\,]$ and $\widetilde J_n(\delta) / \delta^{3/2}$ strictly decreasing.

\paragraph{Step 4 - Rate balance} By the definition of $\mathcal{F}^{\mathrm{KAN}}_r$, $J_n \asymp n^{1/(2r+1)}$. By (C5), $\eta_n = O(n^{-2r/(2r+1)} \log n)$. Set $\delta_n = C_\delta n^{-r/(2r+1)} \sqrt{\log n}$ with $C_\delta$ large enough (depending on $M_{\mathrm{ratio}}$, $C_h(M')$, $C_{\mathrm{app}}$, $K$, and the (C5) asymptotic constants) that, for $n \geq n_*$, the following conditions hold:
\begin{enumerate}
    \item $\delta_n \ge 64\, M_{\mathrm{ratio}}\, H(p_n^{\circ}, p_0)$. By \eqref{eq:rate-approx}: 
    \begin{equation}
        \label{eq:rate_balance_1}
        H(p_n^{\circ}, p_0) \lesssim J_n^{-r} \lesssim n^{-r/(2r+1)}
    \end{equation}
    \item $2\sqrt{M_{\mathrm{ratio}}}\, \widetilde J_n(\delta_n) \leq \sqrt n \delta_n^2$. By \eqref{eq:rate-bracket} with $J_n \lesssim n^{1/(2r+1)}$ and $\log(J_n/\delta_n) + 1 \leq 2 \log n$:
    \begin{equation}
        \label{eq:rate_balance_2}
        \widetilde J_n(\delta_n) \lesssim n^{1/(2(2r+1))} \sqrt{\log n} \delta_n
    \end{equation}
    matching $\sqrt{n} \delta_n^2 = C_\delta n^{1/(2(2r+1))} \sqrt{\log n} \delta_n$.
    \item \begin{equation}
        \label{eq:rate_balance_3}
        \eta_n \leq \delta_n^2: \eta_n \lesssim n^{-2r/(2r+1)} \log n, \qquad \delta_n^2 = C_\delta^2\, n^{-2r/(2r+1)} \log n
    \end{equation}
\end{enumerate} 

\paragraph{Step 5 - Thm. 3.4.12 verification.} Steps 1-4 supply the required conditions: (i) bounded ratio by Step 2; (ii) bracketing integral by Step 3 and Step 4 condition 2 ($\widetilde J_n(\delta_n) \le \sqrt n\, \delta_n^2$ via $M_{\mathrm{ratio}} \ge 1$, hence $\mathcal{J}_{[\,]}(\delta_n, \mathcal{P}_n, H) \le \sqrt n \delta_n^2$); and (iii) discrepancy threshold by Step 4 condition 1 ($\delta_n \ge 64\, M_{\mathrm{ratio}}\, H(p_n^{\circ}, p_0) \geq H(p_n^{\circ}, p_0)$).

\paragraph{Step 6 - Rate transfer} Thm. 3.4.12 is derived from \cite[Thm. 3.4.1]{vandervaartWeakConvergenceEmpirical2023} via \cite[Lem. 3.4.10]{vandervaartWeakConvergenceEmpirical2023}, which supplies the latter theorem's curvature and modulus hypotheses from our bounded ratio (\cref{prop:bounded_likelihood_ratio}) and bracketing integral (\cref{prop:bracketing_integral_bound}). The comparison-discrepancy hypothesis is made trivial by the matched choice $\theta_n = \theta_{n,0} = p_n^{\circ}$. The only modification to accommodate the near-maximiser slack is to verify Thm. 3.4.1's slack tolerance condition. 

Thm. 3.4.1 includes a near-maximiser slack-tolerance that allows the near-max of the criterion to fall short of the comparison-point value $\mathbb{M}_n(\theta_n)$ by $O_P(\delta_n^2)$:
\begin{equation*}
    \mathbb{M}_n(\hat{\theta}_n) \geq \mathbb{M}_n(\theta_n) - O_P(\delta^2_n)
\end{equation*}
where $\mathbb{M}_n(p) = \mathbb{P}_n m_{n, p}$ for $m_{n, p} = \log\frac{p + p_n^{\circ}}{2 p_n^{\circ}}$, $\hat\theta_n = \hat p_n$, $\theta_n = p_n^{\circ}$. 

Concavity of $\log$ gives $m_{n, \hat p_n} \ge \frac{1}{2}\log(\hat p_n/p_n^{\circ})$. \cref{eq:log-likelihood} and assumption (C5) yield:
\begin{equation*}
    \mathbb{P}_n \log(\hat p_n/p_n^{\circ}) = \mathcal{L}_n(\hat g_n) - \mathcal{L}_n(\bar g_n)\geq -\eta_n
\end{equation*}
Hence $\mathbb{P}_n m_{n, \hat p_n} \geq -\eta_n/2 \geq -\delta_n^2/2$ via \cref{eq:rate_balance_3}. Note  $\mathbb{M}_n(\theta_n) = \mathbb{P}_n m_{n, p_n^{\circ}} \equiv 0$ since $m_{n, p_n^{\circ}} = \log 1 = 0$. Thus we have Thm. 3.4.1's slack tolerance at scale $\delta_n^2/2$. 

Thm. 3.4.1 yields $H(\hat p_n, p_n^{\circ}) = O_P^{*}(\delta_n)$. Triangle inequality $H(\hat p_n, p_0) \le H(\hat p_n, p_n^{\circ}) + H(p_n^{\circ}, p_0)$ with \cref{eq:rate_balance_1} and collapsing $O_P^{*}$ to $O_P$ via \cref{lem:measurable_selector} as mentioned:
\begin{equation*}
    H(\hat p_n, p_0) = O_P(\delta_n) = O_P\!\left(n^{-r/(2r+1)} \sqrt{\log n}\right)
\end{equation*}
Since $H(p_g, p_{g^{*}}) = d_H(g, g^{*})$ per \cref{eq:setup-dH},  $d_H(\hat g_n, g^{*}) = O_P(n^{-r/(2r+1)} \sqrt{\log n})$. 

\subsection{Proofs of propositions}
\label{app:proposition_proofs}
\paragraph{Proof of \cref{prop:approximation}} Denote the target's representation $g^* = \Phi^*_D \circ \cdots \circ \Phi^*_1$ with scalar edges $\varphi^*_{\ell, i, j} : D_\ell \to \mathbb{R}$ in the $W^{r, \infty}$-ball of radius $M$, on closed and bounded edge domains $D_\ell$ and layer domains $\mathcal{Z}_\ell = D_\ell^{d_{\ell-1}}$. Construct the approximant $\bar g_n$, with edges defined as B-splines on the slightly enlarged domain $\tilde D_\ell \supseteq D_\ell$ (corresponding $\tilde{\mathcal{Z}}_\ell \supseteq \mathcal{Z}_\ell$). We derive the stated bound and the conditions under which $\bar g_n \in \mathcal{G}_n$. 

To approximate each "true" edge $\varphi^*_{\ell, i, j}$ of $g^{*}$, apply Stein extension to $\varphi^*_{\ell,i,j} \in W^{r,\infty}(D_\ell)$, obtaining a $W^{r,\infty}(\mathbb{R})$ intermediate function with norm inflation $C_{\mathrm{ext}}(r, |D_\ell|)$. Then, restrict to $\tilde D_\ell$ and apply Schumaker's order-$m$ quasi-interpolant \cite[Thm. 6.18]{schumakerSplineFunctionsBasic2007} with $J_n$ uniform knots, yielding B-splines $\tilde\varphi_{\ell, i, j}$. It follows from \cite[Cor. 6.21]{schumakerSplineFunctionsBasic2007} that:
\begin{equation}
    \label{eq:prop1_spline_sup_norm}
    \|\tilde\varphi_{\ell, i, j} - \varphi^*_{\ell, i, j}\|_{\infty, D_\ell} \leq \eta J_n^{-r}
\end{equation}
Further, the derivative $\|\tilde\varphi'_{\ell, i, j}\|_\infty$ is bounded by $L_{\mathrm{edge}}$ (\cite[Thm. 6.25]{schumakerSplineFunctionsBasic2007}). It follows that each $\tilde\varphi_{\ell, i, j}$ is $L_\mathrm{edge}$-Lipschitz on $\tilde{\mathcal{Z}}_\ell$. Finally, $\|\boldsymbol{\theta}^{\tilde\varphi_{\ell, i, j}}\|_\infty \leq C_\theta$ for constant $C_\theta$ if $J_n$ exceeds a threshold $J_{\min}$ (\cite[Thm. 4.42]{schumakerSplineFunctionsBasic2007}). 

Assemble the representation layers $\tilde{\Phi}_\ell$. Each layer coordinate consists of $d_{\ell-1} \leq \max(d_0, Q)$ edges. Per the edges' Lipschitz property, the summation $\tilde{\Phi}_\ell$ is $L_{\mathrm{edge}}\max(d_0, Q)$-Lipschitz on $\tilde{\mathcal{Z}}_\ell$. Further, summing \cref{eq:prop1_spline_sup_norm} over $d_{\ell-1}$ gives $\|\tilde{\Phi}_\ell(\mathbf{u}) - \Phi^*_\ell(\mathbf{u})\|_\infty \leq \eta J_n^{-r} \max(d_0, Q)$ for every $\mathbf{u} \in \mathcal{Z}_\ell$. By \cref{lem:spline_bounds}(i) at coefficient cap $C_\theta$ and summation over $d_{\ell-1}$ inputs, $\tilde{\Phi}_\ell$ maps $\tilde{\mathcal{Z}}_\ell$ into $\tilde{\mathcal{Z}}_{\ell+1}$ and the composition $\tilde{\Phi}_D \circ \cdots \circ \tilde{\Phi}_1$ is well-defined.

Finally, assemble the approximant $\bar{g}_n = \tilde{\Phi}_D \circ \cdots \circ \tilde{\Phi}_1$. Invoking \cref{lem:telescoping} with $L_0 = L_{\mathrm{edge}}\max(d_0, Q), \eta_\ell = \eta J_n^{-r} \max(d_0, Q)$: 
\begin{equation}
    \label{eq:prop1_proof_bound}
    \|\bar g_n - g^*\|_{\infty, \mathcal{Z}} \leq D L_{\mathrm{edge}}^{D-1} \eta \max(d_0, Q)^D J^{-r}_n =: C_{\mathrm{app}} J^{-r}_n
\end{equation}

It remains to enforce $\bar{g}_n \in \mathcal{G}_n$. From the assumption $\|g^*\|_{\infty, \mathcal{Z}} \leq M_0$ and \cref{eq:prop1_proof_bound} it follows $\|\bar{g}_n\|_\infty \leq M_0+C_{\mathrm{app}} J^{-r}_n$. Thus define constant:
\begin{equation*}
    J_0 = \max \left(J_{\min}, \left\lceil(C_{\mathrm{app}}/M_0)^{1/r}\right\rceil\right)
\end{equation*}
Then for $J_n \geq J_0$, $C_{\mathrm{app}} J^{-r}_n \leq M_0$ thus $\|\bar{g}_n\|_\infty \leq 2M_0 = M'$. Coefficient bound $\|\boldsymbol{\theta}^{\tilde\varphi_{\ell, i, j}}\|_\infty \leq C_\theta$ holds since $J_0 \geq J_{\min}$, so each edge of $\bar g_n$ satisfies $\mathcal{G}_n$'s coefficient cap. Thus, $\bar{g}_n \in \mathcal{G}_n$. 

\paragraph{Proof of \cref{prop:entropy}} Apply \cref{lem:spline_bounds} per edge yielding $\|\varphi_{\ell, i, j}^{(\boldsymbol{\theta})}\|_{\infty, I_\ell} \leq C_\theta$, and since edges are piecewise polynomial on the per-edge interval $I_\ell$, with \cref{lem:spline_bounds} (ii) they are Lipschitz of constant $\leq c_{\mathrm{slope}}(J_n+1)C_\theta$. Aggregate edges into each node of layer $\Phi_\ell$, first obtaining per-layer Lipschitz constant:
\begin{equation}
    \label{eq:prop2_ln_bound}
    L_n^{(\boldsymbol{\theta})} \leq c_1(J_n+1)
\end{equation}
for $c_1 = \max(d_0, Q) c_{\mathrm{slope}} C_\theta \geq 1$. Then linearity of B-splines and \cref{lem:spline_bounds} (i) on $\varphi^{(\boldsymbol{\theta} - \boldsymbol{\theta}')}_{\ell, i, j}$ gives per-edge perturbation for every $\mathbf{u} \in \tilde{\mathcal{Z}}_\ell$. Aggregating: 
\begin{equation*}
    \|\Phi_\ell^{(\boldsymbol{\theta})}(\mathbf{u}) - \Phi_\ell^{(\boldsymbol{\theta}')}(\mathbf{u})\|_\infty \leq \max(d_0, Q) \|\boldsymbol{\theta} - \boldsymbol{\theta}'\|_\infty
\end{equation*}
Applying \cref{lem:telescoping}, each layer's perturbation telescopes through the Lipschitz property of the layers: 
\begin{equation}
    \label{eq:composition_lipschitz}
    \|g_{\boldsymbol{\theta}} - g_{\boldsymbol{\theta}'}\|_{\infty, \mathcal{Z}} \leq D\left(L_n^{(\boldsymbol{\theta})}\right)^{D-1} \max(d_0, Q) \|\boldsymbol{\theta} - \boldsymbol{\theta}'\|_\infty = L_n \|\boldsymbol{\theta} - \boldsymbol{\theta}'\|_\infty
\end{equation}
Finally bound $L_n$ in terms of $p_n$. Since $J_n+m \leq p_n$, \cref{eq:prop2_ln_bound} implies $L_n^{(\boldsymbol{\theta})} \leq c_1 p_n$, substituting into the definition of $L_n$: 
\begin{equation*}
    L_n \leq  C_L p_n^{D-1}
\end{equation*}
for an architecture-dependent $C_L \geq 1$. 

Convert bracketing in $\mathcal{G}_n$ to covering in $\Theta_n$ via \cite[Thm. 2.7.17]{vandervaartWeakConvergenceEmpirical2023}. From \cref{eq:composition_lipschitz} we have measurable envelope $L_n$ on $\mathcal{Z}$. For every $\rho > 0$: 
\begin{equation*}
    N_{[\,]}(2 \rho L_n, \mathcal{G}_n, \|\cdot\|_{\infty, \mathcal{Z}}) \leq N(\rho, \Theta_n, \|\cdot\|_\infty)
\end{equation*}
The right side is the $\rho$-covering number of parameter box $\Theta_n$ in $\ell^{\infty}$. Partitioning the axes into $1 + \lfloor C_\theta/\rho
  \rfloor$ intervals of width $2\rho$ gives:
\begin{equation*}
    N(\rho, \Theta_n, \|\cdot\|_\infty) \leq (1 + C_\theta / \rho)^{p_n}
\end{equation*}

For $\varepsilon = 2 \rho L_n \in (0, 2M']$, taking logs:
\begin{equation}
    \label{eq:prop2_entropy_bound}
    \log N_{[\,]}(\varepsilon, \mathcal{G}_n, \|\cdot\|_{\infty, \mathcal{Z}}) \leq p_n \log (1 + 2C_\theta L_n / \varepsilon)
\end{equation}
Let $A = 2C_\theta C_L $ and $y = Ap_n^{D-1}/\varepsilon$, then:
\begin{equation}
    \label{eq:prop2_log_bound}
    \log(1 + 2C_\theta L_n/\varepsilon) \leq \log(1 + Ap_n^{D-1}/\varepsilon) = \log(1 + y) 
\end{equation}
Define the constants:
\begin{equation*}
    p_{\min} = \max\left\{2 e M', \left(2M'\right)^{D-1}\right\}, \qquad
    C_{\mathrm{ent}} = D + \log(2A)
\end{equation*}
For $p_n \geq p_{\min}, \varepsilon \in (0, 2M']$:
\begin{align*}
    p_n/\varepsilon \geq 2eM'/(2M') = e &\implies \log(p_n/\varepsilon) \geq 1 \\ 
    p_n \geq (2M')^{D-1} \geq \varepsilon^{D-1} &\implies (D-1) \log p_n - \log \varepsilon \leq D \log(p_n/\varepsilon)
\end{align*}

Splitting on $y$, if $y < 1$ then $\log(1+y) \leq \log 2 \leq \log(2A)$, and if $y \geq 1$: 
\begin{equation*}
    \log(1+y) \leq \log(2y) = \log(2A) + (D-1)\log p_n - \log \varepsilon \leq \log(2A) + D \log(p_n / \varepsilon) 
\end{equation*}
In both cases and using $\log (p_n /\varepsilon) \geq 1, \log(2A) \geq 0$:
\begin{equation*}
    \log(1+y) \leq (\log(2A) + D) \log(p_n / \varepsilon) = C_{\mathrm{ent}}\log(p_n / \varepsilon)
\end{equation*}
Then via \cref{eq:prop2_entropy_bound,eq:prop2_log_bound}: 
\begin{equation}
    \log N_{[\,]}(\varepsilon, \mathcal{G}_n, \|\cdot\|_{\infty, \mathcal{Z}}) \leq p_n \log(1 + 2C_\theta L_n / \varepsilon) \leq p_n C_{\mathrm{ent}} \log (p_n /\varepsilon) 
\end{equation}

\paragraph{Proof of \cref{prop:bounded_likelihood_ratio}} First observe that the zero set is shared between any $g_1, g_2$ as it depends only on the censoring mechanism: 
\begin{align*}
    \{(\mathbf{x}, y, \Delta) : p_{g}(\mathbf{x}, y, \Delta) = 0\} &= \{(\mathbf{x}, y, 1) : S_C(y \mid \mathbf{x}) = 0\} \\
    &\cup \{(\mathbf{x}, y, 0) : y \in [0, 1),\ f_C(y \mid \mathbf{x}) = 0\} \\
    &\cup \{(\mathbf{x}, 1, 0) : \alpha(\mathbf{x}) = 0\}
\end{align*}
Fix the convention that on this zero set, $\frac{p_{g_1}}{p_{g_2}} = 1 \implies \log \frac{p_{g_1}}{p_{g_2}} = 0$, and focus on the remaining space outside of it, where $S_C, f_C, \alpha > 0$. 
Let $\eta = \|g_1 - g_2\|_{\infty, \mathcal{Z}}$. By MVT on $u \mapsto e^u$ and integrating over $y$ given $g_i(\mathbf{x},y) \in [-B,B]$:
\begin{equation*}
    |\Lambda_{g_1} - \Lambda_{g_2}| \leq e^B\eta
\end{equation*}
Across the three strata of $\mu$: 
\begin{enumerate}[label=(\roman*)]
    \item $\Delta = 1$:
\begin{align*}
    \frac{p_{g_1}}{p_{g_2}} &= \frac{S_C e^{g_1} e^{-\Lambda_{g_1}}}{S_C e^{g_2} e^{-\Lambda_{g_2}}} = e^{g_1 - g_2}e^{-(\Lambda_{g_1} - \Lambda_{g_2})}\\
    \implies \left| \log \frac{p_{g_1}}{p_{g_2}}\right| &\leq |g_1 - g_2| + |\Lambda_{g_1} - \Lambda_{g_2}| \leq \eta + e^B\eta \leq (1+e^B)\eta 
\end{align*}
    \item $\Delta = 0, y \in [0, 1)$:
    \begin{equation*}
        \frac{p_{g_1}}{p_{g_2}} = \frac{f_C\,e^{-\Lambda_{g_1}}}{f_C \, e^{-\Lambda_{g_2}}} \implies \left|\log \frac{p_{g_1}}{p_{g_2}}\right| \leq |\Lambda_{g_1} - \Lambda_{g_2}| \leq e^B\eta \leq (1+e^B)\eta 
    \end{equation*}
    \item $\Delta = 0, y = 1$: Same as the above, with $f_C \rightarrow \alpha, y \rightarrow 1$.
\end{enumerate}
Thus the required bound holds across the $\mu$-support.

\paragraph{Proof of \cref{prop:bracketing_integral_bound}} The bracketing integral is:
\begin{equation*}
    \mathcal{J}_{[\,]}\left(\delta,\mathcal{P}_n,H\right)
    = \int_0^\delta \sqrt{1 + \log N_{[\,]}\left(\varepsilon,\mathcal{P}_n,H\right)}d\varepsilon.
\end{equation*}
Recall the sup-norm bound $\mathcal{G}_n \subseteq \{g : \|g\|_{\infty, \mathcal{Z}} \le M'\}$. Invoke \cref{cor:sup_norm_to_hellinger_bracket_count} at $B = M'$ to transfer the Hellinger count to sup-norm count.
\begin{equation*}
    \log N_{[\,]}(\varepsilon, \mathcal{P}_n, H) \leq \log N_{[\,]} \left(\varepsilon / \sqrt{C_h(M')}, \mathcal{G}_n, \|\cdot\|_{\infty, \mathcal{Z}}\right), \qquad C_h(B) := \frac{5}{4} e^{3B}
\end{equation*}
Pass to \cref{prop:entropy} at scale $\varepsilon / \sqrt{C_h(M')}$:
\begin{equation*}
    \log N_{[\,]}(\varepsilon, \mathcal{P}_n, H) \leq C_{\mathrm{ent}}p_n(\log(p_n / \varepsilon) + \kappa), \qquad \kappa := \tfrac{1}{2}\log C_h(M')
\end{equation*}
on $\varepsilon \in (0, 2 M' \sqrt{C_h(M')}]$ (so the substituted scale lies in $(0, 2 M']$) and $J_n \ge J^*_{\mathrm{ent}} := \min\{J \in \mathbb{N}_{\ge 1} : E(J + m) \ge p_{\min}\}$ (so $p_n \ge p_{\min}$).

Set $c_0 := E(1 + m)$. $J_n \ge 1$ gives $p_n = E J_n + E m \le c_0J_n$, and a threshold ensuring $\log(J_n/\varepsilon) \ge 0$ on the operating domain is $J_n \ge \lceil 2 M' \sqrt{C_h(M')}\rceil$. Define
\begin{equation*}
    J^* := \max\left(J^*_{\mathrm{ent}},\lceil 2 M' \sqrt{C_h(M')}\rceil\right).
\end{equation*}

For $J_n \ge J^*$ and $\varepsilon \in (0, 2 M' \sqrt{C_h(M')}]$, $\log(J_n/\varepsilon) \ge 0$. The inequality $A + B \le \max(1, B)(A + 1)$ for $A \ge 0$, $B \in \mathbb{R}$ at $A = \log(J_n/\varepsilon)$, $B = \log c_0 + \kappa$ gives $\log(p_n/\varepsilon) + \kappa \le \max(1, \log c_0 + \kappa)[\log(J_n/\varepsilon) + 1]$. Substituting $p_n \le c_0 J_n$ and setting $C_0 := C_{\mathrm{ent}}c_0\max(1, \log c_0 + \kappa)$:
\begin{equation}
    \label{eq:bra-cons}
    \log N_{[\,]}(\varepsilon, \mathcal{P}_n, H) \leq C_0J_n(\log(J_n/\varepsilon) + 1)
\end{equation}
$J_n[\log(J_n/\varepsilon) + 1] \ge 1$ on this domain, so the integrand in the bracketing integral satisfies:
\begin{align*}
    \sqrt{1 + \log N_{[\,]}\left(\varepsilon,\mathcal{P}_n,H\right)}
        &\leq \sqrt{1 + C_0J_n(\log(J_n/\varepsilon) + 1)} \\
        &\leq \sqrt{(C_0 + 1)J_n}\sqrt{\log(J_n/\varepsilon) + 1}
\end{align*}
$\sqrt{(C_0 + 1)J_n}$ is constant in $\varepsilon$ and factors out. Substituting $u = \log(J_n/\varepsilon)$ and integrating by parts:
\begin{equation*}
    \int_0^\delta \sqrt{\log (J_n/\varepsilon) + 1} d\varepsilon \leq 2\delta\sqrt{\log (J_n/\delta) + 1}
\end{equation*}
Therefore, the bracketing integral becomes: 
\begin{equation*}
    \mathcal{J}_{[\,]}\left(\delta,\mathcal{P}_n,H\right) \leq \sqrt{(C_0 + 1)J_n} \cdot 2\delta\sqrt{\log(J_n / \delta) + 1}
\end{equation*}
Substitute $K = 2\sqrt{C_0+1}$:
\begin{equation}
    \label{eq:bracketing_integral_bound_jn}
    \mathcal{J}_{[\,]}\left(\delta,\mathcal{P}_n,H\right) \leq K\sqrt{J_n}\delta\sqrt{\log(J_n / \delta) + 1} =: \widetilde J_n(\delta)
\end{equation}
for $\delta \in (0, 2 M' \sqrt{C_h(M')}]$ and $J_n \ge J^*$. $M_0 \ge 1 \implies 2 M' \sqrt{C_h(M')} > \sqrt 2$, so $(0, \sqrt 2] \subseteq (0, 2 M' \sqrt{C_h(M')}]$ and thus \cref{eq:bracketing_integral_bound_jn} holds for $\delta \in (0, \sqrt 2]$, giving result (iii). 

It remains to verify the shape of $\widetilde J_n(\delta)$ on $(0, \sqrt 2]$. Set $u(\delta) := \log(J_n/\delta)$. $J_n \ge J^* \ge 32 > \sqrt 2 \ge \delta$ gives $u > 0$. Differentiating $\delta\sqrt{u + 1}$ in $\delta$ (with $u'(\delta) = -1/\delta$) gives $\frac{2 u + 1}{2 \sqrt{u + 1}} > 0$, so $\widetilde J_n = K \sqrt{J_n}\delta\sqrt{u + 1}$ is strictly increasing, giving result (i).

Finally, $\widetilde J_n(\delta) / \delta^{3/2} = K \sqrt{J_n}\delta^{-1/2}\sqrt{u + 1}$. $\delta^{-1/2}$ and $\sqrt{u + 1}$ are strictly positive and strictly decreasing on $\delta \in (0, \sqrt 2]$, so their product is strictly decreasing, giving result (ii). 

\subsection{Technical Lemmas} 
\begin{lemma}[Telescoping]
    \label{lem:telescoping}
    Let pairs $(\Phi_\ell, \tilde\Phi_\ell)_{\ell = 1}^D$ of continuous maps on nested domains $\mathcal{Z}_\ell \subseteq \tilde{\mathcal{Z}}_\ell$ ($\ell \in \{1, \ldots, D + 1\}$) with $\Phi_\ell : \mathcal{Z}_\ell \to \mathcal{Z}_{\ell + 1}$ and $\tilde\Phi_\ell : \tilde{\mathcal{Z}}_\ell \to \tilde{\mathcal{Z}}_{\ell + 1}$, where:
    \begin{enumerate}
        \item Each $\tilde\Phi_\ell$ is $L_0$-Lipschitz on $\tilde{\mathcal{Z}}_\ell$ in the sup-norm, $L_0 \ge 1$, and
        \item $\|\tilde\Phi_\ell(\mathbf{u}) - \Phi_\ell(\mathbf{u})\|_\infty \le \eta_\ell$ for every $\mathbf{u} \in \mathcal{Z}_\ell$.
    \end{enumerate}
    Denote the compositions $F_D = \Phi_D \circ \cdots \circ \Phi_1$, $\tilde F_D = \tilde\Phi_D \circ \cdots \circ \tilde\Phi_1$. 
    Then, the composition error:
    \begin{equation}
          \|F_D - \tilde F_D\|_{\infty, \mathcal{Z}_1} \leq D\, L_0^{D - 1}\, \max_{1 \le \ell \le D} \eta_\ell
    \end{equation}
\end{lemma}
Proof: This is shown by telescoping $F_D - \tilde F_D$ into $D$ incremental one-layer swaps of $\Phi_\ell$ for $\tilde \Phi_\ell$. Sum the post-swap maps' cumulative Lipschitz constants $L_0^{D-\ell}$ from assumption (1) and the per-layer perturbations $\eta_\ell$ from assumption (2). The final inequality follows from $L_0^{D - \ell} \leq L_0^{D - 1}$.

\begin{lemma}[Spline bounds]
    \label{lem:spline_bounds}
    For an order-$m$ B-spline expansion $s_{\boldsymbol{\theta}} = \sum_k \theta_k B^{I, J}_k$ on a closed bounded interval $I$ with $J$ uniform interior knots, $m$-fold boundary knots, and mesh $\bar \Delta = |I|/(J+1)$ it holds that:
    \begin{enumerate}[label=(\roman*)]
        \item $\|s_{\boldsymbol{\theta}}\|_{\infty, I} \leq \|\boldsymbol{\theta}\|_\infty$, and 
        \item $\|s'_{\boldsymbol{\theta}}\|_{L^{\infty}(I)} \leq c_{\mathrm{slope}} \|\boldsymbol{\theta}\|_\infty / \bar\Delta$ for $c_{\mathrm{slope}}$ depending only on $m$. 
    \end{enumerate}
\end{lemma}
Proof: Partition of unity on $I$ \cite[Thm. 4.20]{schumakerSplineFunctionsBasic2007} and non-negativity of $B_k$ give property (i). For property (ii), express the right derivative $D_+ s$ as an order $m-1$ B-spline expansion, whose coefficients are bounded by $c_{\mathrm{slope}} \|\boldsymbol{\theta}\|_\infty / \bar \Delta$ \cite[Thm. 5.9]{schumakerSplineFunctionsBasic2007}, and convert this to a bound on the derivative at every point on $I$ except the knot points, obtaining the essential supremum $\|s'_{\boldsymbol{\theta}}\|_{L^{\infty}(I)}$. 

\begin{lemma}[Sup-norm Hellinger bound]
    \label{lem:sup_norm_to_hellinger}
    Let $B \in [0, \infty)$ and define $C_h(B) = \frac{5}{4}e^{3B}$. For every pair of measurable log-hazards $g_1, g_2 : \mathcal{Z} \to \mathbb{R}$ with $\|g_i\|_{\infty, \mathcal{Z}} \leq B$ ($i = 1, 2$):
    \begin{equation}
        H^2(p_{g_1}, p_{g_2}) \leq C_h(B)\|g_1 - g_2\|^2_{\infty, \mathcal{Z}}
    \end{equation}
\end{lemma}
Proof: Decompose $H^2(p_{g_1}, p_{g_2}) = H^2_{\mathrm{ac}}(p_{g_1}, p_{g_2}) + H^2_{\mathrm{at}}(p_{g_1}, p_{g_2})$ via $\mu = \mu_{\mathrm{ac}} + \mu_{\mathrm{at}}$ 
\begin{align*}
    H^2_{\mathrm{ac}}(p_{g_1}, p_{g_2}) &= \sum_{\Delta \in \{0, 1\}} \int_{\mathcal{X}} \int_0^1 \left(\sqrt{p_{g_1}(\mathbf{x}, y, \Delta)} - \sqrt{p_{g_2}(\mathbf{x}, y, \Delta)}\right)^2\, dy\, dP_X(\mathbf{x}) \\
    H^2_{\mathrm{at}}(p_{g_1}, p_{g_2})
    &= \int_{\mathcal{X}} \left(\sqrt{p_{g_1}(\mathbf{x}, 1, 0)} - \sqrt{p_{g_2}(\mathbf{x}, 1, 0)}\right)^2\, dP_X(\mathbf{x})
\end{align*}

Define $a_g(\mathbf{x}, y) = e^{g(\mathbf{x}, y)/2}$
and $b_g(y \mid \mathbf{x}) = e^{-\Lambda_g(y \mid \mathbf{x})/2}$ where $\Lambda_g$ is the cumulative hazard induced by $g$. Factorise $\sqrt{p_g}$ over the three possibilities:
\begin{equation*}
    \begin{cases}
        \sqrt{p_g(\mathbf{x}, y, 1)} = \sqrt{S_C(y \mid \mathbf{x})}a_g(\mathbf{x},y)b_g(y \mid \mathbf{x}) & \Delta = 1, y \in [0, 1) \\
        \sqrt{p_g(\mathbf{x}, y, 0)} = \sqrt{f_C(y \mid \mathbf{x})}\, b_g(y \mid \mathbf{x}) & \Delta = 0, y \in [0, 1) \\
        \sqrt{p_g(\mathbf{x}, 1, 0)} = \sqrt{\alpha(\mathbf{x})}\, b_g(1 \mid \mathbf{x}) & \Delta = 0, y = 1
    \end{cases}
\end{equation*}

Let $\eta = \|g_1 - g_2\|_{\infty, \mathcal{Z}}$ and derive Lipschitz bounds: 
\begin{enumerate}[label=(\roman*)]
    \item \begin{equation}
        \label{eq:hellinger-a-lip}
        |a_{g_1}(\mathbf{x}, y) - a_{g_2}(\mathbf{x}, y)| \leq \frac{1}{2} e^{B/2} \eta
    \end{equation}
    by MVT on $u \mapsto e^{u/2}$ given $g_i(\mathbf{x}, y) \in [-B,B]$.
    \item 
    \begin{equation}
        \label{eq:hellinger-ell-lip}
        |\Lambda_{g_1}(y \mid \mathbf{x}) - \Lambda_{g_2}(y \mid \mathbf{x})| \leq e^B \eta
    \end{equation}
    by MVT on $u \mapsto e^u$ and integration over $y$.
    \item 
    \begin{equation}
        \label{eq:hellinger-b-lip}
        |b_{g_1}(y \mid \mathbf{x}) - b_{g_2}(y \mid \mathbf{x})| \leq \frac{1}{2} e^B \eta 
    \end{equation}
    by MVT on $u \mapsto e^{-u/2}$ and (ii).
\end{enumerate}

Per-factor upper bounds: $a_{g_i} \le e^{B/2}$ from $g_i \le B$; $b_{g_i} \le 1$ from $\Lambda_{g_i} \ge 0$; $S_C \le 1$ from (C3). Product Lipschitz from $a_1 b_1 - a_2 b_2 = (a_1 - a_2)\, b_1 + a_2\, (b_1 - b_2)$ with (i), (iii):
\begin{equation}
    \label{eq:hellinger-ab-lip}
    |a_{g_1} b_{g_1} - a_{g_2} b_{g_2}| \leq \frac{1}{2} e^{B/2}(1 + e^B)\, \eta \leq e^{3 B / 2}\, \eta 
\end{equation}

Substitute the squared difference and integrate against $\mu$:
\begin{enumerate}[label=(\roman*)]
    \item $\Delta = 1$, $y \in [0, 1)$:
    \begin{align*}
        & \left(\sqrt{p_{g_1}(\mathbf{x}, y, 1)} - \sqrt{p_{g_2}(\mathbf{x}, y, 1)}\right)^2
            = S_C(y \mid \mathbf{x})\, (a_{g_1} b_{g_1} - a_{g_2} b_{g_2})^2
            \leq S_C(y \mid \mathbf{x}) e^{3 B} \eta^2 \\
        \implies\ & \int_{\mathcal{X}} \int_0^1 \left(\sqrt{p_{g_1}(\mathbf{x}, y, 1)} - \sqrt{p_{g_2}(\mathbf{x}, y, 1)}\right)^2\, dy\, dP_X(\mathbf{x}) \leq e^{3 B} \eta^2,
    \end{align*}
    using $\int_0^1 S_C(y \mid \mathbf{x})\, dy \le 1$ from $S_C \le 1$ and $P_X(\mathcal{X}) = 1$.

    \item $\Delta = 0$, $y \in [0, 1)$:
    \begin{align*}
        & \left(\sqrt{p_{g_1}(\mathbf{x}, y, 0)} - \sqrt{p_{g_2}(\mathbf{x}, y, 0)}\right)^2
            = f_C(y \mid \mathbf{x})\, (b_{g_1} - b_{g_2})^2
            \leq f_C(y \mid \mathbf{x})\, \frac{1}{4} e^{2 B} \eta^2 \\
        \implies\ & \int_{\mathcal{X}} \int_0^1  \left(\sqrt{p_{g_1}(\mathbf{x}, y, 0)} - \sqrt{p_{g_2}(\mathbf{x}, y, 0)}\right)^2\, dy\, dP_X(\mathbf{x}) \leq \frac{1}{4} e^{2 B} \eta^2 \int_{\mathcal{X}} \int_0^1 f_C(y \mid \mathbf{x})\, dy\, dP_X(\mathbf{x})
    \end{align*}

    \item $\Delta = 0$, $y = 1$:
    \begin{align*}
        & \left(\sqrt{p_{g_1}(\mathbf{x}, 1, 0)} - \sqrt{p_{g_2}(\mathbf{x}, 1, 0)}\right)^2
            = \alpha(\mathbf{x})\, (b_{g_1}(1 \mid \mathbf{x}) - b_{g_2}(1 \mid \mathbf{x}))^2
            \leq \alpha(\mathbf{x})\, \frac{1}{4} e^{2 B}\, \eta^2 \\
        \implies\ & \int_{\mathcal{X}} \left(\sqrt{p_{g_1}(\mathbf{x}, 1, 0)} - \sqrt{p_{g_2}(\mathbf{x}, 1, 0)}\right)^2\, dP_X(\mathbf{x}) \leq \frac{1}{4} e^{2 B}\, \eta^2 \int_{\mathcal{X}} \alpha(\mathbf{x})\, dP_X(\mathbf{x})
    \end{align*}
\end{enumerate}

Combine (ii) and (iii) via (C3) $\int_0^1 f_C(y \mid \mathbf{x})\, dy + \alpha(\mathbf{x}) = 1$ and $P_X(\mathcal{X}) = 1$:
\begin{equation*}
    \frac{1}{4} e^{2 B}\, \eta^2 \int_{\mathcal{X}} \left( \int_0^1 f_C(y \mid \mathbf{x})\, dy + \alpha(\mathbf{x}) \right) dP_X(\mathbf{x}) = \frac{1}{4} e^{2 B}\, \eta^2
\end{equation*}

Summing the three strata into $H^2 = H^2_{\mathrm{ac}} + H^2_{\mathrm{at}}$ and since $B \ge 0$:
\begin{equation*}
    H^2(p_{g_1}, p_{g_2}) \leq e^{3 B}\, \eta^2 + \frac{1}{4} e^{2 B}\, \eta^2 \leq \frac{5}{4} e^{3 B}\, \eta^2 = C_h(B)\, \|g_1 - g_2\|^2_{\infty, \mathcal{Z}}
\end{equation*}

\begin{corollary}
    \label{cor:sup_norm_to_hellinger_bracket_count}
    Under \cref{lem:sup_norm_to_hellinger}, for density class $\mathcal{P} = \{p_g: g \in \mathcal{G}\}$ with $\mathcal{G} \subseteq \{g : \|g\|_{\infty, \mathcal{Z}} \le B\}$ and every $\varepsilon > 0$:
    \begin{equation}
        \log N_{[\,]}(\varepsilon, \mathcal{P}, H) \leq \log N_{[\,]}(\varepsilon / \sqrt{C_h(B)}, \mathcal{G}, \|\cdot\|_{\infty, \mathcal{Z}})
    \end{equation}
\end{corollary}
Proof: Apply \cite[Thm. 2.7.17]{vandervaartWeakConvergenceEmpirical2023} to class $\mathcal{F} = \{\sqrt{p_g} : g \in \mathcal{G}\}$ indexed by $(\mathcal{G}, \|\cdot\|_{\infty, \mathcal{Z}})$ with norm $L^2(\mu)$. The 3-case Lipschitz bounds on $|\sqrt{p_{g_1}} - \sqrt{p_{g_2}}|$ from \cref{lem:sup_norm_to_hellinger}'s factorisation \cref{eq:hellinger-a-lip,eq:hellinger-ell-lip,eq:hellinger-b-lip} give envelope:
\begin{equation*}
    F(\mathbf{x}, y, \Delta) =
    \begin{cases}
        e^{3 B / 2} & \Delta = 1, y \in [0, 1) \\
        \frac{1}{2}\, e^B\, \sqrt{f_C(y \mid \mathbf{x})} & \Delta = 0, y \in [0, 1) \\
        \frac{1}{2}\, e^B\, \sqrt{\alpha(\mathbf{x})} & \Delta = 0, y = 1
    \end{cases}
\end{equation*}
Integrating $F^2$ over $\mu$ and combining the two $\Delta = 0$ strata $\int_0^1 f_C(y \mid \mathbf{x})\, dy + \alpha(\mathbf{x}) = 1$:
\begin{equation*}
    \|F\|^2_{L^2(\mu)} \leq e^{3 B} + \frac{1}{4} e^{2 B} \leq \frac{5}{4} e^{3 B} = C_h(B)
\end{equation*}
\cite[Thm. 2.7.17]{vandervaartWeakConvergenceEmpirical2023} then gives, for every $\eta > 0$ and under $N(\eta, \mathcal{G}, \|\cdot\|_{\infty, \mathcal{Z}}) < \infty$:
\begin{equation}
    \label{eq:kosorok-invoke}
    N_{[\,]}\left(2 \sqrt{C_h(B)} \eta, \{\sqrt{p_g} : g \in \mathcal{G}\}, L^2(\mu)\right) \leq N\left(\eta, \mathcal{G}, \|\cdot\|_{\infty, \mathcal{Z}}\right)
\end{equation}
An $L^2(\mu)$-bracket $(\psi_L, \psi_U)$ for $\sqrt{p_g}$ produces a Hellinger bracket $(\ell, u)$ for $p_g$ of the same width via $\ell = (\psi_L \vee 0)^2$, $u = (\psi_U \vee 0)^2$. Then $\ell \leq p_g \leq u$ $\mu$ almost everywhere (since $\sqrt{p_g} \geq 0$), and $H^2(u, \ell) \leq \int (\psi_U - \psi_L)^2\, d\mu$ since $\sqrt{u} = \psi_U \vee 0$, $\sqrt{\ell} = \psi_L \vee 0$, and $x \mapsto x \vee 0$ is 1-Lipschitz. Thus:
\begin{equation}
    \label{eq:hell-L2-corr}
    N_{[\,]}(\varepsilon, \mathcal{P}, H) \leq N_{[\,]}(\varepsilon, \{\sqrt{p_g} : g \in \mathcal{G}\}, L^2(\mu))
\end{equation}
Midpoints of a sup-norm $r$-bracket cover of $\mathcal{G}$ form an $r/2$-cover:
\begin{equation}
    \label{eq:cov-bra}
    N(r / 2, \mathcal{G}, \|\cdot\|_{\infty, \mathcal{Z}}) \leq N_{[\,]}(r, \mathcal{G}, \|\cdot\|_{\infty, \mathcal{Z}})
\end{equation}
Substitute $\eta = \varepsilon / (2 \sqrt{C_h(B)})$ in \cref{eq:kosorok-invoke} and $r = \varepsilon / \sqrt{C_h(B)}$ in \cref{eq:cov-bra}. Then through \cref{eq:hell-L2-corr}:
\begin{align*}
    \log N_{[\,]}(\varepsilon, \mathcal{P}, H)
        &\leq \log N_{[\,]}(\varepsilon, \{\sqrt{p_g} : g \in \mathcal{G}\}, L^2(\mu)) \\
        &\leq \log N(\varepsilon / (2 \sqrt{C_h(B)}), \mathcal{G}, \|\cdot\|_{\infty, \mathcal{Z}}) \\
        &\leq \log N_{[\,]}(\varepsilon / \sqrt{C_h(B)}, \mathcal{G}, \|\cdot\|_{\infty, \mathcal{Z}})
\end{align*}

\begin{lemma}[Measurable selector]
    \label{lem:measurable_selector}
    For every $\eta_n \ge 0$ there exists a Borel-measurable $\hat g_n : \Omega \to \mathcal{G}_n$ that is an $\eta_n$-near-maximiser of $\mathcal{L}_n$ (i.e., $\mathcal{L}_n(\hat g_n) \ge \sup_{g \in \mathcal{G}_n} \mathcal{L}_n(g) - \eta_n$).
\end{lemma}
$\mathcal{G}_n$ is parameterised by its B-spline coefficient vector $\boldsymbol{\theta} \in \Theta_n^{\mathrm{feas}} \subset \mathbb{R}^{p_n}$ where $\Theta_n^{\mathrm{feas}} = \{\boldsymbol{\theta} \in \Theta_n : \|g_{\boldsymbol{\theta}}\|_{\infty, \mathcal{Z}} \leq M'\}$. The map $(\boldsymbol{\theta}, \omega) \mapsto \mathcal{L}_n(g_{\boldsymbol{\theta}}; \omega)$ is Carath\'eodory because it is (a) continuous in $\boldsymbol{\theta}$ for each $\omega$ via B-spline linearity in $\boldsymbol{\theta}$ and dominated convergence with envelope $e^{M'}$ on the cumulative-hazard integral $\int_0^{Y_i} e^{g_{\boldsymbol{\theta}}(X_i, s)}\, ds$; and (b) Borel in $\omega$ for each $\boldsymbol{\theta}$. This implies joint Borel measurability and gives $\sup_{\boldsymbol{\theta}} \mathcal{L}_n(g_{\boldsymbol{\theta}}; \omega) = \sup_{\boldsymbol{\theta} \in \Theta_n^{\circ}} \mathcal{L}_n(g_{\boldsymbol{\theta}}; \omega)$ for countable dense $\Theta_n^{\circ} \subset \Theta_n^{\mathrm{feas}}$, hence Borel in $\omega$. Define the set:
\begin{equation*}
      \mathcal{S}_n(\omega) = \left\{\boldsymbol{\theta} \in \Theta_n^{\mathrm{feas}} : \mathcal{L}_n(g_{\boldsymbol{\theta}}; \omega) \ge \sup\nolimits_{\Theta_n^{\mathrm{feas}}} \mathcal{L}_n(\cdot; \omega) - \eta_n\right\}
\end{equation*}
This set is non-empty, closed-valued, and has Borel graph. Kuratowski-Ryll-Nardzewski \cite[Thm. 17.13]{aliprantisInfiniteDimensionalAnalysis1999} then supplies a Borel selector $\hat{\boldsymbol{\theta}}_n : \Omega \to \Theta_n^{\mathrm{feas}}$ with $\hat{\boldsymbol{\theta}}_n(\omega) \in \mathcal{S}_n(\omega)$. The composition $\hat g_n := g_{\hat{\boldsymbol{\theta}}_n}$ is thus Borel via the continuous parameterisation $\boldsymbol{\theta} \mapsto g_{\boldsymbol{\theta}}$.

\section{Implementation Details}
\label{app:implementation_details}
\ac{KAPLAN-HR} is implemented with the \texttt{pykan} library v0.2.8 by \citet{liuKANKolmogorovarnoldNetworks2025} for the \ac{KAN} layers. Each edge is a B-spline plus a SiLU residual. We use the library's default parameters throughout, which we list in \cref{tab:pykan_defaults}.

During training, the knot vectors are periodically re-fitted to the observed activation distribution via \texttt{pykan}'s grid-extension at a tunable schedule of epochs $\mathcal{U}$. Parameters are optimised with Adam at a tunable constant learning rate using mini-batches of tunable size $B$. Training stops when the validation \ac{NLL} has not decreased below its running minimum for $P$ consecutive epochs. The complete training procedure is given in \cref{alg:training}. 

We use right-Riemann integration for both prediction and training. The prediction grid $\boldsymbol{\tau}$ used for evaluation (and validation during training) is initialised at $K+1$ quantiles of the training observed times $\{Y_i\}_{i=1}^n$ (with $\tau_0 \equiv 0$ and duplicates removed). During training, computation of the \ac{NLL} requires $\hat{\Lambda}_\text{HR}(Y_i \mid \mathbf{x}_i)$ for each subject. For efficiency, we evaluate this on a sub-grid $\boldsymbol{\tau}_\mathcal{B}$ of bin indices $\{b_i := \max\{k : \tau_k \leq Y_i\} : i \in \mathcal{B}\}$ in the mini-batch $\mathcal{B}$. We apply an $L_2$ shrink penalty to the predicted log-hazard in each batch: 
\begin{equation*}
    \mathcal{L}_{\text{shrink}} = \lambda_{\text{shrink}} \cdot \frac{1}{| \mathcal{B}||\boldsymbol{\tau}_{\mathcal{B}}|}\sum_{i\in\mathcal{B}}\sum_{\tau  
  \in \boldsymbol\tau_\mathcal{B}} \hat g(\mathbf{x}_i, \tau)^2
\end{equation*}

\paragraph{Factored first-layer evaluation}
\label{sec:kan_l1_factorisation}
Evaluating $\mathrm{KAN}_{\boldsymbol{\theta}}(\mathbf{x}_i, t)$ at $n \times K$ (sample, time) pairs is a major cost for \ac{KAPLAN-HR}. This cost can be reduced by exploiting the additive structure of \cref{eq:kan_layer}, decomposing the first layer output as:
\begin{equation}
    z_j^{(1)} = \sum_{i=1}^{d} \varphi_{1, j,i}(x^{(i)}) + \varphi_{1, j,d+1}(t) =: F_j(\mathbf{x}) + G_j(t).
    \label{eq:kan_l1_factorisation}
\end{equation}
Since $F_j$ depends only on $\mathbf{x}$ and $G_j$ only on $t$, they may be evaluated independently and then combined by broadcast addition. Subsequent layers operate on the full $z^{(1)}$, which mixes $\mathbf{x}$ and $t$, so the factorisation does not extend beyond the first layer. 

\begin{table}[htbp]
\centering
\small
\caption{KAN configuration. Parameter names follow the \texttt{pykan} v0.2.8 API. `tuned' entries are swept per (dataset, variant)..}
\label{tab:pykan_defaults}
\begin{tabular}{@{}l c c l@{}}
\toprule
Parameter & Simulations & Real Data & Description \\
\midrule
\texttt{k} & $3$ & tuned $\in\{2,3,4\}$ & Spline degree (our $m$)\\
\texttt{grid} & adaptive & tuned $\in\{2,3,4,5\}$ & Grid intervals (our $J_n$)\\
\texttt{grid\_range} & $[-1,1]$ & $[-1,1]$ & Initial knot range \\
\texttt{grid\_eps} & $0.02$ & $0.02$ & $0$: quantile knots; $1$: uniform \\
\texttt{noise\_scale} & $0.3$ & $0.3$ & Coefficient init noise \\
\texttt{scale\_sp} & $1$ & $1$ & $w_s$ init scale \\
\texttt{scale\_base\_mu} & $0$ & $0$ & $w_b$ init mean \\
\texttt{scale\_base\_sigma} & $1$ & $1$ & $w_b$ init sigma \\
\texttt{sp\_trainable} & \texttt{True} & \texttt{True} & $w_s$ trained \\
\texttt{sb\_trainable} & \texttt{True} & \texttt{True} & $w_b$ trained \\
\texttt{affine\_trainable} & \texttt{False} & \texttt{False} & Affine frozen \\
\texttt{base\_fun} & \texttt{silu} & \texttt{silu} & Residual function \\
\midrule
Grid re-fits & $1$ (at init) & tuned $\in\{5,10,15,20\}$ & \texttt{update\_grid} calls per fit \\
Patience ($P$) & 50 & 50 & early stopping patience\\
\bottomrule
\end{tabular}
\end{table}

\begin{algorithm}[htbp]
\caption{Training \ac{KAPLAN-HR}}\label{alg:training}
\begin{algorithmic}[1]
\Require Data $\mathcal{D} = \{(\mathbf{x}_i, Y_i, \Delta_i)\}_{i=1}^n$, time grid $\boldsymbol{\tau} = (\tau_0 \equiv 0, \tau_1, \ldots, \tau_K)$, batch size $B$, grid update epochs $\mathcal{U}$, patience $P$
\State Initialise $\mathrm{KAN}_{\boldsymbol{\theta}}$ with input dimension $d+1$
\State Compute bin indices $b_i \gets \max\{k : \tau_k \le Y_i\}$ for $i = 1, \ldots, n$
\For{epoch $e = 1, 2, \ldots$}
    \If{$e \in \mathcal{U}$} re-fit B-spline knot vectors
    \EndIf
    \For{each batch $\mathcal{B} \subset \{1, \ldots, n\}$, $|\mathcal{B}| = B$}
        \State $(a_1, \ldots, a_R) \gets \mathrm{sorted}\!\left(\mathrm{unique}(\{b_i : i \in \mathcal{B}\})\right)$ \Comment{$R$ active bin indices}
        \State Sub-grid $\boldsymbol{\tau}_\mathcal{B} \gets (\tau_{\mathcal{B}, r} := \tau_{a_r})_{r=1}^{R}$ with widths $\tilde{\delta}_r \gets \tau_{\mathcal{B}, r} - \tau_{\mathcal{B}, {r-1}}$ ($\tau_{\mathcal{B}, 0} \equiv 0$)
        \State Evaluate $g_{i,r} \gets \mathrm{KAN}_{\boldsymbol{\theta}}(\mathbf{x}_i, \tau_{\mathcal{B},r})$ for $i \in \mathcal{B}$, $r = 1, \ldots, R$ \Comment{Factored, cf.\ \cref{eq:kan_l1_factorisation}}
        \State $h_{i,r} \gets \exp(g_{i,r})$
        \State $H_{i,r} \gets \sum_{m=1}^{r} h_{im}\,\tilde{\delta}_m$ \Comment{Cumulative sum (right-Riemann)}
        \State $r_i \gets$ position of $b_i$ in $(a_1, \ldots, a_R)$
        \State $\mathcal{L}_\text{HR} \gets \tfrac{1}{B}\sum_{i \in \mathcal{B}} \left[H_{i, r_i} - \Delta_i\, g_{i, r_i}\right] + \mathcal{L}_{\mathrm{shrink}}$
        \State Adam step on $\nabla_{\boldsymbol{\theta}} \mathcal{L}_\text{HR}$
    \EndFor
    \State \textbf{if} validation \ac{NLL} has not improved for $P$ epochs \textbf{then break}
\EndFor
\State \Return $\mathrm{KAN}_{\boldsymbol{\theta}}$
\end{algorithmic}
\end{algorithm}

\section{Experimental Details}
\label{app:experimental_details}
\subsection{Computing Platform} All experiments were run on the Cambridge Service for Data-Driven Discovery (CSD3) high-performance computing platform, using Intel Ice Lake processors. We set the memory limit to 64GB. 

\subsection{Simulation Experiments}
\label{app:experimental_details_simulation}
We define four synthetic \acp{DGP}:
\begin{align*}
    &\text{DGP-1:} \; g^*(\mathbf{x},t) = \sin(2\pi x_1) + 0.5\cos(2\pi x_2) - 0.3\sin(\pi t) + 0.2 \\
    &\text{DGP-2:} \; g^*(\mathbf{x},t) = |2 x_1 - 1| + 0.5\cos(2\pi x_2) - 0.3\sin(\pi t) + 0.2 \\
    &\text{DGP-3:} \; g^*(\mathbf{x},t) = \sin\left(\pi(x_1 + 0.5x_2)\right) - 0.3t + 0.2 \\
    &\text{DGP-4:} \; g^*(\mathbf{x}, t) = x_1^2 \cdot t + 0.3 \sin(\pi x_2) - 0.4 
\end{align*}
\acp{DGP} 1,3, and 4 are analytic, thus the smoothness exponent $r$ for rate analysis is capped by the spline order. We use cubic B-splines for all simulation experiments, so $r=4$. \ac{DGP}-2 contains the tent term $|2 x_1 - 1|$ which admits only one weak derivative, yielding $r=1$. 

For each \ac{DGP} we generate $R=100$ independent training sets plus a shared testing set of $2000$ observations. We further split each training set $80/20$ and use the validation sub-set for early stopping. We draw i.i.d. covariates $x_1, \ldots, x_5 \sim \mathrm{Uniform}(0,1)$ ($x_3, x_4, x_5$ are intentionally left as noise) and generate continuous event times by inversion from the conditional hazard $\lambda (t \mid x) = \exp(g(x, t))$ on $t \in (0,10]$. Ground-truth survival $S(t \mid x) = \exp\left(-\int_0^t\exp(g(x, u))du\right)$ is computed on a 200-point time grid by trapezoidal integration. Censoring times $C \sim \mathrm{Exp}(\lambda_c)$ are drawn independently of $T$, and we apply administrative censoring at horizon $t = 3$ and calibrate $\lambda_c$ to yield an overall censoring rate of $\sim 30\%$. 

For \acp{DGP} 1 and 2 we fit \ac{KAPLAN-HR} models with structure $[d_0, 1]$, and for \acp{DGP} 3 and 4 we use deeper structure $[d_0, 3, 1]$. The spline grid is set adaptively to $J_n = \lceil n^{1/(2r+1)} \rceil$. The other model parameters are listed in \cref{tab:pykan_defaults}. The \ac{GAM} baselines are fitted using the \texttt{mgcv} R package \cite{woodSmoothingParameterModel2016}, setting $k=10$ for the covariates and $k=15$ for the time and the remaining parameters. 

\subsection{Real-World Clinical Dataset Experiments}
\label{app:experimental_details_benchmark}

\paragraph{Datasets and Preprocessing} We evaluate on six standard survival analysis datasets: \ac{METABRIC} \cite{curtisGenomicTranscriptomicArchitecture2012a}, \ac{RotGBSG} \cite{foekensUrokinaseSystemPlasminogen2000, schumacherRandomized221994}, \ac{NWTCO} \cite{breslowDesignAnalysisTwoPhase1999}, \ac{FLCHAIN} \cite{dispenzieriUseNonclonalSerum2012}, and \ac{SUPPORT} \cite{knausSUPPORTPrognosticModel1995}, with covariates selected to match \citet{kvammeContinuousDiscretetimeSurvival2021}; and \ac{MIMIC-III} \cite{johnsonMIMICIIIClinicalDatabase2016}, following the preprocessing of \citet{purushothamBenchmarkingDeepLearning2018}. Each dataset is split 10 times stratified on the event indicator into 64\% train, 16\% validation, and 20\% test sets with random seeds 0-9. Covariates and time are standardised using training set statistics at each split. Event times are min-max normalised to $[0,1]$ prior to splitting. Rows with any missing values are dropped. 

\paragraph{Baseline Models} We compare performance against: Cox proportional hazards, implemented by \texttt{lifelines} \cite{davidson-pilonLifelinesSurvivalAnalysis2019}; a generalised additive model (GAM), implemented by the \texttt{mgcv} R package \cite{woodSmoothingParameterModel2016}; HARE \cite{kooperbergHazardRegression1995}, implemented by the \texttt{polspline} R package; DeepSurv \cite{katzmanDeepSurvPersonalizedTreatment2018}, CoxTime \cite{kvammeTimetoeventPredictionNeural2019}, and DeepHit \cite{leeDeepHitDeepLearning2018}, implemented by \texttt{pycox} \cite{kvammeTimetoeventPredictionNeural2019}; Deep Survival Machines (DSM) \cite{nagpalDeepSurvivalMachines2021}, implemented by \texttt{auton-survival} \cite{nagpalAutonsurvivalOpenSourcePackage}; CoxKAN \cite{knottenbeltCoxKANKolmogorovArnoldNetworks2025}; and SuMo-Net \cite{rindtSurvivalRegressionProper2022}.  

\paragraph{Hyperparameter Tuning} All tunable models undergo Bayesian hyperparameter optimisation using Weights \& Biases \cite{weightsandbiasesSweepsTuneHyperparameters2026} to maximise the time-dependent concordance index (C-TD) \cite{antoliniTimedependentDiscriminationIndex2005} on the validation set in split 0 (out of 10), separately for each dataset, over 50 trials per (dataset, model) pair. \cref{tab:appendix_hyperparameter_sweeps} gives the sweep ranges for each model hyperparameter. 

\paragraph{Scoring} Survival predictions are evaluated on: the time-dependent concordance index (C-TD) by \citet{antoliniTimedependentDiscriminationIndex2005}, implemented by \texttt{pycox} \cite{kvammeTimetoeventPredictionNeural2019}; the integrated Brier score (IBS) \cite{grafAssessmentComparisonPrognostic1999}, implemented by \texttt{scikit-survival} \cite{polsterlScikitsurvivalLibraryTimetoevent2020}; the integrated calibration index (ICI) \cite{austinGraphicalCalibrationCurves2020}, implemented by \texttt{lifelines} \cite{davidson-pilonLifelinesSurvivalAnalysis2019}, measured at the median training set event time (per split); and the D-calibration p-value (D-CAL) \cite{haiderEffectiveWaysBuild2020}, implemented by \texttt{SurvivalEVAL} \cite{qiSurvivalEVALComprehensiveOpensource2024}. All metrics except D-CAL are averaged over the 10 data splits and reported as the mean $\pm$ standard deviation. The D-CAL is reported as the number of splits (out of 10) with $p > 0.05$. Statistical comparisons between \ac{KAPLAN-HR} and the best-scoring baseline model on each dataset use the two-sided Wilcoxon signed-rank test on the 10 per-split metric values, with Holm-Bonferroni correction applied across the six datasets per metric. 

\begin{table}[htbp]
\centering
\caption{Hyperparameter search spaces for all models.}
\label{tab:appendix_hyperparameter_sweeps}
\small
\begin{tabular}{@{}lll@{}}
\toprule
Hyperparameter & Range / values & Distribution \\
\midrule
\multicolumn{3}{@{}l}{\textbf{KAPLAN-HR}} \\
\midrule
Learning rate            & $[10^{-4},\, 10^{-1}]$                             & log-uniform \\
Spline grid $J_n$          & $\{2,\, 3,\, 4,\, 5\}$                             & categorical \\
Spline order $m$         & $\{2,\, 3,\, 4\}$                                  & categorical \\
Layers $D$ & $\{1,\, 2,\, 3\}$                                       & categorical \\
Hidden dimension $Q$     & $[2, 32]$ & uniform \\
Grid update frequency    & $\{5,\, 10,\, 15,\, 20\}$                          & categorical \\
Shrink penalty $\lambda_{\mathrm{shrink}}$ & $[0,\, 10^{-3}]$                        & uniform \\
Batch size     & $128$ & constant \\
Max training epochs & $200$ & constant \\
Shared time grid size $L$ & $128$ & constant \\ 
\midrule
\multicolumn{3}{@{}l}{\textbf{DeepSurv, CoxTime, DeepHit}} \\
\midrule
Hidden layers            & $\{1, 2, 4\}$                                      & categorical \\
Nodes per layer          & $\{32, 64, 128, 256\}$                             & categorical \\
Learning rate            & $[10^{-4.5},\, 10^{-1.5}]$                         & log-uniform \\
Weight decay             & $\{0,\, 0.01,\, 0.02,\, 0.05,\, 0.1,\, 0.2,\, 0.4\}$ & categorical \\
Dropout                  & $[0,\, 0.5]$                                       & uniform \\
Batch normalisation      & $\{\text{true},\, \text{false}\}$                  & categorical \\
Batch size               & $\{32,\, 64,\, 128,\, 256,\, 512\}$                & categorical \\
\midrule
\multicolumn{3}{@{}l}{\textbf{CoxTime only}} \\
\midrule
Shrinkage $\lambda$      & $\{0,\, 0.001,\, 0.01,\, 0.1\}$                    & categorical \\
Log-duration transform   & $\{\text{true},\, \text{false}\}$                  & categorical \\
\midrule
\multicolumn{3}{@{}l}{\textbf{DeepHit only}} \\
\midrule
Ranking weight $\alpha$  & $[0,\, 1]$                                         & uniform \\
Ranking scale $\sigma$   & $\{0.1,\, 0.25,\, 1,\, 5,\, 10\}$                  & categorical \\
Num.\ durations          & $\{10,\, 25,\, 50,\, 100\}$                        & categorical \\
\midrule
\multicolumn{3}{@{}l}{\textbf{Deep Survival Machines (DSM)}} \\
\midrule
Num.\ experts $k$        & $\{4,\, 6,\, 8\}$                                  & categorical \\
Discount factor          & $\{0.5,\, 0.75,\, 1.0\}$                           & categorical \\
Hidden layers            & $\{1,\, 2\}$                                       & categorical \\
Nodes per layer          & $\{50,\, 100\}$                                    & categorical \\
Output distribution      & $\{\text{LogNormal},\, \text{Weibull}\}$           & categorical \\
Learning rate            & $\{10^{-3},\, 10^{-4}\}$                           & categorical \\
Batch size & $100$ & constant \\
\midrule
\multicolumn{3}{@{}l}{\textbf{SuMo-Net}} \\
\midrule
Covariate sub-net depth  & $\{1,\, 2,\, 4\}$                                  & categorical \\
Time sub-net depth       & $\{1,\, 2,\, 4\}$                                  & categorical \\
Nodes per layer          & $\{8,\, 16,\, 32\}$                                & categorical \\
Dropout                  & $\{0,\, 0.1,\, 0.2,\, 0.3,\, 0.4,\, 0.5\}$         & categorical \\
Weight decay             & $\{0,\, 0.01,\, 0.02,\, 0.05,\, 0.1,\, 0.2,\, 0.4\}$ & categorical \\
Batch size               & $\{25,\, 50,\, 100,\, 250\}$                       & categorical \\
Learning rate     & $10^{-3}$                                          & constant \\
\midrule
\multicolumn{3}{@{}l}{\textbf{CoxKAN}} \\
\midrule
\multicolumn{3}{@{}p{0.95\linewidth}@{}}{Same ranges used in the authors' reproducibility code \cite{knottenbeltCoxKANKolmogorovArnoldNetworks2025}.} \\
\bottomrule
\end{tabular}
\end{table}

\section{Real-World Clinical Dataset Results}
\label{app:benchmark_results}
\cref{tab:appendix_hyperparameter_results} reports the tuned hyperparameters for \ac{KAPLAN-HR}. \cref{tab:supp-ctd-hr,tab:supp-ibs-hr,tab:supp-ici-hr,tab:supp-dcal-hr} present the complete benchmark results. 
\begin{table}[tbp]
\centering
\caption{Tuned hyperparameters for KAPLAN-HR per dataset (best run from Bayesian sweep). LR=Adam learning rate; $J_n$ =spline knot count; $m$ = spline order; $D$ = number of layers; $Q$ = hidden layer dimension; Grid upd. = frequency of spline grid updating; $\lambda_{\mathrm{shrink}}$ = shrink penalty. }
\label{tab:appendix_hyperparameter_results}
\small
\begin{tabular}{@{}lrrrrrrr@{}}
\toprule
Dataset & LR & $J_n$ & $m$ & $D$ & $Q$ & Grid upd. & $\lambda_{\mathrm{shrink}}$ \\
\midrule
METABRIC & 0.000161 & 4 & 3 & 3 & 24 & 5 & $4.61\times 10^{-4}$ \\
MIMIC-III & 0.00665 & 2 & 4 & 3 & 6 & 20 & $1.94\times 10^{-4}$ \\
RotGBSG & 0.00179 & 5 & 4 & 1 & 6 & 5 & $3.30\times 10^{-4}$ \\
FLCHAIN & 0.00243 & 2 & 2 & 3 & 2 & 20 & $1.98\times 10^{-4}$ \\
NWTCO & 0.00186 & 4 & 4 & 2 & 4 & 10 & $4.84\times 10^{-4}$ \\
SUPPORT & 0.000119 & 3 & 2 & 3 & 22 & 10 & $3.25\times 10^{-4}$ \\
\bottomrule
\end{tabular}
\end{table}
\begin{table}[htbp]
\centering
\caption{Time-dependent concordance (C-TD, $\uparrow$, mean $\pm$ std) benchmark results.}
\label{tab:supp-ctd-hr}
\scriptsize
\begin{tabular}{@{}lrrrrrr@{}}
\toprule
Model & METABRIC & MIMIC-III & RotGBSG & FLCHAIN & NWTCO & SUPPORT \\
\midrule
CoxPH & $.6460 \pm .0183$ & $.7375 \pm .0106$ & $.6374 \pm .0222$ & $.7093 \pm .1619$ & $.7120 \pm .0229$ & $.6609 \pm .0111$ \\
DeepSurv & $.6582 \pm .0173$ & $.8005 \pm .0086$ & $.6322 \pm .0271$ & $.7908 \pm .0128$ & $.7154 \pm .0307$ & $.6656 \pm .0105$ \\
CoxTime & $.6729 \pm .0159$ & $.7976 \pm .0122$ & $.6336 \pm .0313$ & $.7899 \pm .0123$ & $.7101 \pm .0200$ & $.6639 \pm .0117$ \\
DeepHit & $.6646 \pm .0158$ & $.8102 \pm .0084$ & $.6458 \pm .0253$ & $.7257 \pm .0913$ & $.7175 \pm .0252$ & $\mathbf{.6732 \pm .0127}$ \\
DSM & $.6695 \pm .0259$ & $.8034 \pm .0101$ & $.6620 \pm .0226$ & $.7895 \pm .0118$ & $.7153 \pm .0158$ & $.6656 \pm .0122$ \\
GAM & $.6345 \pm .0096$ & $.6420 \pm .0707$ & $.6389 \pm .0376$ & $.6663 \pm .0287$ & $.6024 \pm .0242$ & $.6458 \pm .0094$ \\
HARE & $.6661 \pm .0152$ & $.7863 \pm .0121$ & $.6520 \pm .0313$ & $.7782 \pm .0535$ & $.7146 \pm .0214$ & $.6641 \pm .0126$ \\
CoxKAN & $.6577 \pm .0167$ & $.8049 \pm .0106$ & $.6472 \pm .0257$ & $.7891 \pm .0119$ & $\mathbf{.7194 \pm .0257}$ & $.6653 \pm .0105$ \\
SuMo-Net & $.6546 \pm .0176$ & $.7617 \pm .0876$ & $.6304 \pm .0492$ & $.7886 \pm .0112$ & $.7180 \pm .0200$ & $.6420 \pm .0222$ \\
KAPLAN-HR & $\mathbf{.6763 \pm .0161}$ & $\mathbf{.8114 \pm .0098}$ & $\mathbf{.6671 \pm .0213}$ & $\mathbf{.7959 \pm .0128}$ & $.7180 \pm .0243$ & $.6706 \pm .0102$ \\
\bottomrule
\end{tabular}
\end{table}

\begin{table}[htbp]
\centering
\caption{Integrated Brier Score (IBS, $\downarrow$, mean $\pm$ std) benchmark results.}
\label{tab:supp-ibs-hr}
\scriptsize
\begin{tabular}{@{}lrrrrrr@{}}
\toprule
Model & METABRIC & MIMIC-III & RotGBSG & FLCHAIN & NWTCO & SUPPORT \\
\midrule
CoxPH & $\mathbf{.1602 \pm .0069}$ & $.2012 \pm .0295$ & $.1755 \pm .0097$ & $.1168 \pm .0332$ & $.1055 \pm .0033$ & $.1844 \pm .0061$ \\
DeepSurv & $.1607 \pm .0074$ & $.1985 \pm .0301$ & $.1773 \pm .0090$ & $.1004 \pm .0029$ & $.1067 \pm .0074$ & $\mathbf{.1828 \pm .0059}$ \\
CoxTime & $.1603 \pm .0066$ & $.1866 \pm .0250$ & $.1804 \pm .0100$ & $.1107 \pm .0129$ & $.1070 \pm .0031$ & $.1832 \pm .0064$ \\
DeepHit & $.1734 \pm .0069$ & $.1850 \pm .0257$ & $.1903 \pm .0162$ & $.3008 \pm .1633$ & $.1075 \pm .0076$ & $.2587 \pm .0071$ \\
DSM & $.1635 \pm .0069$ & $.1841 \pm .0264$ & $.1764 \pm .0080$ & $.0981 \pm .0028$ & $\mathbf{.0983 \pm .0036}$ & $.1851 \pm .0062$ \\
GAM & $.1794 \pm .0077$ & $.2015 \pm .0293$ & $.1854 \pm .0278$ & $.1197 \pm .0033$ & $.1096 \pm .0025$ & $.2006 \pm .0062$ \\
HARE & $.1644 \pm .0069$ & $.1897 \pm .0320$ & $.1781 \pm .0094$ & $.1294 \pm .0956$ & $.1016 \pm .0066$ & $.1857 \pm .0071$ \\
CoxKAN & $.1651 \pm .0088$ & $.1959 \pm .0278$ & $.1749 \pm .0108$ & $.1127 \pm .0020$ & $.1045 \pm .0037$ & $.1837 \pm .0064$ \\
SuMo-Net & $.1647 \pm .0056$ & $.2445 \pm .0871$ & $.1869 \pm .0199$ & $.1126 \pm .0050$ & $.1022 \pm .0031$ & $.1988 \pm .0070$ \\
KAPLAN-HR & $.1658 \pm .0080$ & $\mathbf{.1588 \pm .0169}$ & $\mathbf{.1723 \pm .0102}$ & $\mathbf{.0980 \pm .0027}$ & $.1021 \pm .0041$ & $.1834 \pm .0066$ \\
\bottomrule
\end{tabular}
\end{table}

\begin{table}[htbp]
\centering
\caption{ICI at median event time (ICI Q2, $\downarrow$, mean $\pm$ std) benchmark results.}
\label{tab:supp-ici-hr}
\scriptsize
\begin{tabular}{@{}lrrrrrr@{}}
\toprule
Model & METABRIC & MIMIC-III & RotGBSG & FLCHAIN & NWTCO & SUPPORT \\
\midrule
CoxPH & $.0246 \pm .0107$ & $.0082 \pm .0039$ & $.0366 \pm .0145$ & $.0479 \pm .0586$ & $.0286 \pm .0067$ & $\mathbf{.0293 \pm .0081}$ \\
DeepSurv & $\mathbf{.0231 \pm .0074}$ & $.0088 \pm .0020$ & $.0364 \pm .0200$ & $.0180 \pm .0061$ & $.0318 \pm .0136$ & $.0381 \pm .0106$ \\
CoxTime & $.0294 \pm .0129$ & $.0224 \pm .0020$ & $.0381 \pm .0201$ & $.0489 \pm .0325$ & $.0181 \pm .0061$ & $.0355 \pm .0097$ \\
DeepHit & $.0742 \pm .0199$ & $.0092 \pm .0013$ & $.0618 \pm .0246$ & $.2916 \pm .2825$ & $.0351 \pm .0095$ & $.2616 \pm .0071$ \\
DSM & $.0311 \pm .0126$ & $.0092 \pm .0029$ & $.0503 \pm .0245$ & $.0145 \pm .0061$ & $.0303 \pm .0093$ & $.0394 \pm .0035$ \\
GAM & $.0460 \pm .0138$ & $.0368 \pm .0245$ & $.0583 \pm .0507$ & $.0244 \pm .0106$ & $\mathbf{.0156 \pm .0153}$ & $.0415 \pm .0080$ \\
HARE & $.0320 \pm .0087$ & $\mathbf{.0079 \pm .0020}$ & $.0449 \pm .0187$ & $.0561 \pm .1219$ & $.0438 \pm .0218$ & $.0305 \pm .0086$ \\
CoxKAN & $.0252 \pm .0092$ & $.0092 \pm .0025$ & $.0348 \pm .0163$ & $.0321 \pm .0253$ & $.0299 \pm .0100$ & $.0323 \pm .0096$ \\
SuMo-Net & $.0416 \pm .0135$ & $.0151 \pm .0156$ & $.0534 \pm .0138$ & $.0214 \pm .0089$ & $.0267 \pm .0091$ & $.0775 \pm .0302$ \\
KAPLAN-HR & $.0350 \pm .0114$ & $.0094 \pm .0017$ & $\mathbf{.0315 \pm .0087}$ & $\mathbf{.0135 \pm .0053}$ & $.0362 \pm .0059$ & $.0433 \pm .0107$ \\
\bottomrule
\end{tabular}
\end{table}

\begin{table}[htbp]
\centering
\caption{D-calibration pass count $(k/n)$ benchmark results. Cells with $k = n$ (all splits pass at $p > 0.05$) in bold.}
\label{tab:supp-dcal-hr}
\scriptsize
\begin{tabular}{@{}lrrrrrr@{}}
\toprule
Model & METABRIC & MIMIC-III & RotGBSG & FLCHAIN & NWTCO & SUPPORT \\
\midrule
CoxPH & \textbf{10/10} & \textbf{10/10} & 9/10 & 8/10 & \textbf{10/10} & 5/10 \\
DeepSurv & \textbf{10/10} & \textbf{10/10} & 9/10 & \textbf{10/10} & \textbf{10/10} & 3/10 \\
CoxTime & \textbf{10/10} & \textbf{10/10} & \textbf{10/10} & 5/10 & \textbf{10/10} & 5/10 \\
DeepHit & 9/10 & \textbf{10/10} & 5/10 & 0/10 & \textbf{10/10} & 0/10 \\
DSM & 9/10 & \textbf{10/10} & 9/10 & \textbf{10/10} & \textbf{10/10} & 0/10 \\
GAM & \textbf{10/10} & 4/10 & 9/10 & \textbf{10/10} & \textbf{10/10} & 0/10 \\
HARE & \textbf{10/10} & \textbf{10/10} & \textbf{10/10} & 9/10 & 8/10 & 9/10 \\
CoxKAN & \textbf{10/10} & \textbf{10/10} & \textbf{10/10} & 5/10 & \textbf{10/10} & 4/10 \\
SuMo-Net & 1/10 & 3/10 & 5/10 & 1/10 & \textbf{10/10} & 0/10 \\
KAPLAN-HR & 9/10 & \textbf{10/10} & \textbf{10/10} & \textbf{10/10} & \textbf{10/10} & 7/10 \\
\bottomrule
\end{tabular}
\end{table}

%%%%%%%%%%%%%%%%%%%%%%%%%%%%%%%%%%%%%%%%%%%%%%%%%%%%%%%%%%%%

%\clearpage
%\input{checklist.tex}

\end{document}